\documentclass{article}

\PassOptionsToPackage{numbers, compress}{natbib}

 \usepackage[main, final]{neurips_2026}

\usepackage[utf8]{inputenc} 
\usepackage[T1]{fontenc}    
\usepackage{hyperref}       
\usepackage{url}            
\usepackage{booktabs}       
\usepackage{amsfonts}       
\usepackage{nicefrac}       
\usepackage{microtype}      
\usepackage{xcolor}         

\usepackage{amsmath}
\usepackage{amssymb}
\usepackage{mathtools}
\usepackage{amsthm}
\usepackage{multirow}
\usepackage[table]{xcolor}
\newcommand{\gr}{\color{gray}} 
\usepackage{makecell}
\usepackage{enumitem}
\usepackage{adjustbox}
\usepackage{varwidth}
\usepackage{caption} 
\usepackage[most]{tcolorbox}
\definecolor{darkgreen}{RGB}{0,120,0}
\definecolor{blue}{RGB}{0,0,220}
\definecolor{darkorange}{RGB}{205,102,0}
\definecolor{red}{RGB}{220,0,0}
\definecolor{darkgreen}{RGB}{34,139,34}
\usepackage{graphicx}
\usepackage{subcaption}
\usepackage{wrapfig}

\newif\ifhighlight
\highlightfalse 
\newcommand{\hl}[1]{%
  \ifhighlight
    \textcolor{blue}{#1}%
  \else
    #1%
  \fi
}

\title{Beyond Prediction: Steering VLM Agents with Retrospective World Modeling}

\author{%
Yongjiang Liu\textsuperscript{1},
Jie Zhang\textsuperscript{1}\thanks{Corresponding author. Primary contact: \texttt{\{yliunr@connect.ust.hk}, \texttt{sejzhang@cse.ust.hk}\}.},
Haoyue Zhang\textsuperscript{1},
Jingcai Guo\textsuperscript{2},
Deze Zeng\textsuperscript{3},
Song Guo\textsuperscript{1}
\\[3pt]
\textsuperscript{1}The Hong Kong University of Science and Technology
\quad
\textsuperscript{2}The Hong Kong Polytechnic University
\\
\textsuperscript{3}China University of Geoscience
}

\begin{document}

\maketitle

\begin{abstract}
Equipping VLM agents with world modeling capabilities has shown strong potential for complex reasoning and long-horizon planning, while reducing the dependence of policy learning on costly real-world interactions. Existing methods mainly rely on prospective simulation to predict the consequences of candidate actions. However, this forward-only paradigm focuses on ``\textit{what will happen next}'' and provides limited constraints for verifying whether an action is causally consistent with the observed state transition, which can lead to plausible-looking but physically incoherent behaviors. In this paper, we challenge the view of world modeling as only prospective prediction and introduce Retrospective World Modeling, a new agent learning paradigm that enables agents to reason backward by estimating the retrospective attribution distribution $P(\hat{a}_{t}|s_t, s_{t+1})$ for the action that most likely caused a given transition. Based on this capability, we formulate the \texttt{S}elf-\texttt{C}onsistency \texttt{R}eward (\texttt{SCR}), an intrinsic signal that measures the probabilistic consistency between the policy action and the retrospective explanation. Integrating SCR into reinforcement learning provides dense transition-level feedback and steers agents toward behaviors that are both task-effective and physically grounded. Extensive experiments across diverse agentic tasks show that our method substantially improves policy robustness and generalization over prospective-only world modeling baselines.
\end{abstract}

\section{Introduction}\label{sec:intro}
\begin{figure}[h]
  \centering
    \includegraphics[width = 1.00\textwidth]{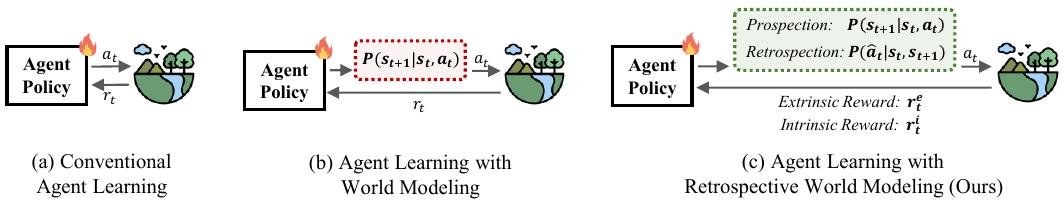}
    \caption{Comparison between our retrospective world modeling framework and conventional agent learning frameworks.}
    \label{fig:comparison_framework}
\end{figure}

Reinforcement Learning (RL) has recently emerged as a pivotal paradigm for advancing Vision–Language Model (VLM) agents, allowing them to transcend static reasoning and engage in complex, multi-turn interactions within dynamic environments~\cite{Guo_2025,openai2024openaio1card,xi2025rise}, ranging from embodied exploration~\cite{shridhar2021alfworld,yang2025embodiedbench} to open-ended digital tasks~\cite{zhou2024webarena,chae2025web}.
Unlike single-turn tasks, success in these scenarios demands not just abstract planning but precise long-horizon decision-making, where every action cascades into future states.
To navigate such dynamics,  
world modeling~\cite{lecun2022path,wang-etal-2025-world,hafner2023mastering,richens2025general} has been introduced as a critical cognitive scaffold for autonomous agents.
As illustrated in Fig.~\ref{fig:comparison_framework}(b), unlike the reactive ``\textit{Reason-Act}'' loop~\cite{yao2023react} (Fig.~\ref{fig:comparison_framework}(a)), contemporary methods~\cite{xing2025critiquesworldmodels,wang2025vagen} adopt a prospective ``\textit{Observe-Predict-Act}'' regime.
By internalizing world knowledge to simulate ``\textit{what will happen next}'', these agents can mentally rehearse actions and anticipate consequences before real-world execution, significantly enhancing policy robustness and effectiveness.

While existing methods have made remarkable progress by internalizing world modeling capabilities directly into the policy to predict future states, we identify two critical limitations in this prospective-only paradigm. 
First, the reasoning process remains essentially open-loop~\cite{zhang2025worldinworldworldmodelsclosedloop,ye2025reinforcementlearninginverserewards}. The agent inherently assumes its predictive foresight is accurate, lacking a mechanism to verify whether its planned actions causally align with the actual environmental dynamics. 
For instance, in a partially observable environment like \textit{Sokoban}, an agent might hallucinate that an \texttt{<Up>} action will move a box upward, ignoring a blocking wall. 
Without real-time verification, the agent continues planning based on this flawed premise, leading to inevitable plan collapse in a dead end~\cite{pflueger2025safetyassessmentreinforcementlearning}. 
Second, this lack of verification represents a missed opportunity for intrinsic feedback.
By overlooking the causal consistency between action and the resulting state transition, current methods discard a rich source of dense supervision. This forces the agent to rely solely on sparse, delayed extrinsic rewards upon task completion, making credit assignment extremely difficult in long-horizon tasks.
This compels us to rethink: \textit{Is a forward-only world modeling sufficient to equip an agent with a robust internal belief that is both causally consistent and self-correcting?}

In this paper, we posit that endowing agents with a bidirectional understanding of environmental dynamics can achieve more robust reasoning and decision-making. Beyond merely functioning as a forward simulator $P(s_{t+1} | s_t, a_t)$\footnote{Here, $P$ is just a conceptual notation for the agent's internal world model of environmental dynamics. In our framework, this capability is instantiated by the unified policy network $\pi_{\theta}$ (Sec.~\ref{sec:method}).} that predicts future outcomes, an internal world model should also serve as a retrospective analyst, capable of deducing the underlying causality $P(\hat{a}_{t} | s_t, s_{t+1})$ behind the observed transitions.
This retrospective capability is pivotal: it complements the agent to verify the logical soundness of its actions against actual environmental dynamics, rather than relying solely on plausible-looking predictions.
To this end, we propose Retrospective World Modeling (RWM), a novel agent learning paradigm that augments agents with the ability to ``\textit{Reason backward}'' (i.e., inferring the action responsible for the observed state change) and further used for policy optimization.
Building upon RWM, we introduce the \texttt{S}elf-\texttt{C}onsistency \texttt{R}eward (\texttt{SCR}), a self-supervised signal that regularizes policy optimization towards actions whose effect can be consistently explained by the observed changes.   
Integrating SCR into RL establishes a closed-loop learning process that complements prospective planning with retrospective verification, shifting intrinsic supervision from curiosity-driven exploration toward exploitation-oriented consistency regularization.

We empirically evaluate our method across diverse agentic tasks, lifting the average success rate from 8\% to 81\% with strong generalization.
Leveraging the granularity of retrospective feedback, our method exhibits higher action effectiveness and rapid convergence compared to baselines.
Subsequent analyses validate the necessity of internalizing this intrinsic signal and assess various reward formulations and context regulation strategies to mitigate reward hacking.

\section{Related Work}\label{sec:related_work}
\textbf{RL for LLM and VLM Agents.}
Recent studies have explored reinforcement learning (RL) to train long-horizon multi-turn LLM and VLM agents.
Diverse capabilities have been boosted, such as 
planning~\cite{zhu2025plannerr1rewardshapingenables,paglieri2025learningplanefficientlyallocating,atasever2025multiagentpathfindingoffline},
tool use~\cite{singh2025agenticreasoningtoolintegration,zheng2025deepeyesincentivizingthinkingimages,wu2025vtoolr1vlmslearnthink},
memory~\cite{zhong2024memorybank,yan2026memoryr1enhancinglargelanguage},
self-improvement~\cite{zuo2025ttrl,huang2026rzeroselfevolvingreasoningllm,zhou2025sweetrltrainingmultiturnllm}, reasoning~\cite{wang2025beyond,yang2025towards},
perception~\cite{tan2025reasonrft,zhu2025shuffler1efficientrlframework}.
RL algorithms, such as PPO~\cite{schulman2017proximalpolicyoptimizationalgorithms}, GRPO~\cite{guo2025deepseek}
and DAPO~\cite{yu2025dapoopensourcellmreinforcement}, etc., form a spectrum from general policy gradients to specialized preference learning.

\textbf{World Modeling.} 
As a cornerstone of model-based RL~\cite{sutton1991dyna,ha2018recurrent,lecun2022path}, world models enable predictive planning and foresight~\cite{guan2023leveraging,qiao2024agent}, achieving remarkable decision-making success via latent dynamics in systems like MuZero~\cite{schrittwieser2020mastering} and the Dreamer family~\cite{hafner2020dream,hafner2025mastering}. 
Recently adapted for LLM/VLM agents, this paradigm explicitly models state transitions (e.g., VAGEN~\cite{wang2025vagen}) or simulates complex domains ranging from web browsing~\cite{gu2025is,feng2025webworldmodels,chae2025web,fang-etal-2025-webevolver}, mathematical reasoning~\cite{hao2023reasoning}, embodied control~\cite{xiang2023language,wang-etal-2025-world}, and games~\cite{wang2025cogitoergoludoagent} to high-fidelity video prediction~\cite{gladstone2025energybasedtransformersscalablelearners,brooks2024video,he2025active}. 
Beyond explicit modeling, some methods internalize strategies~\cite{zhang2025agent,chen2025internalizing}, leveraging early exploration to align internal planning with environmental feedback.

\textbf{Inverse Dynamics for Curiosity and Action Grounding.}
Inverse dynamics models infer actions from state transitions, and have been used in two closely related lines of work. In classical RL, curiosity-driven methods use inverse dynamics to learn action-relevant representations and compute forward prediction errors as intrinsic rewards for exploration~\cite{pathak2017curiosity,burda2018largescale,hong2020adversarial}. In embodied and robotic learning, recent world-action models and video-based policy learning frameworks use inverse dynamics or action-conditioned world modeling to recover executable actions from visual transitions or generated future rollouts, such as PIDM~\cite{tian2025predictive}, Motus~\cite{bi2025motus}, DreamZero~\cite{ye2026world}, GigaWorld-0/Policy~\cite{team2025gigaworld,ye2026gigaworld}. These methods mainly employ inverse dynamics as an auxiliary representation-learning objective, an exploration signal, or an action decoding module. In contrast, RWM repurposes inverse action inference as a retrospective consistency validator for VLM-agent policy learning. Rather than relying on an external inverse model or rewarding prediction error to encourage exploration, RWM performs cross-turn retrospective attribution within the VLM reasoning process and converts discriminative action-transition consistency into an exploration-oriented dense intrinsic reward for directly regularizing RL policy optimization.

\textbf{Summary.} 
Despite these variations, internal world modeling in agents remains overwhelmingly prospective, tasked exclusively with answering ``\textit{what will happen next?}''. 
We argue that this unidirectional perspective is fundamentally incomplete. Instead, we propose RWM, endowing agents with a bidirectional understanding of dynamics that unifies forward planning with retrospective causal verification for robust decision-making.

\section{Preliminary}\label{sec:preliminary}
\subsection{Problem formulation}
We formulate the multi-turn interaction of VLM agents as a Partially Observable Markov Decision Process (POMDP), defined by the tuple $(\mathcal{S}, \mathcal{O}, \mathcal{A}, \mathcal{T}, \mathcal{R}, \gamma)$. 
Here, $\mathcal{S}$ denotes the latent state space of the environment, which is typically complex and not fully observable. $\mathcal{O}$ represents the observation space (e.g., visual inputs and textual feedback), and $\mathcal{A}$ is the action space (e.g., API calls or high-level commands). The transition function $\mathcal{T}: \mathcal{S} \times \mathcal{A} \rightarrow \mathcal{S} $ governs the environment's dynamics, while the reward function $\mathcal{R}$ evaluates task progress given a specific goal $g$.

At each turn $t$, the agent cannot access the true state $s_t \in \mathcal{S}$. Instead, it receives a partial observation $o_t \sim \mathcal{O}(s_t)$ and selects an action $a_t$ based on the interaction history $h_t = (o_1, a_1, \dots, o_{t-1}, a_{t-1}, o_t)$. This decision process is governed by a policy $\pi_\theta$ parameterized by a VLM, i.e., $a_t \sim\pi_\theta(\cdot|h_t)$.
Upon executing $a_t$, the environment transitions to a new state $s_{t+1}$
and emits a scalar reward $r_t$.
The agent's objective is to learn an optimal policy $\pi_\theta$ that maximizes the expected cumulative return:
\begin{equation}
\max_{\theta} \; \mathbb{E}_{\pi_\theta} \left[ \sum_{t=1}^{T} \gamma^{t-1} r_t \right],
\end{equation}
where $\gamma$ denotes the discount factor. The interaction continues until a terminal condition is met or a maximum horizon is reached, yielding a sequential decision-making loop.

\subsection{Prospective Internal World Modeling in VLM Agents}
\label{sec:conventional_wm}

While the POMDP formulation defines the learning objective for multi-turn interaction, a purely reactive policy $\pi_\theta(a_t|h_t)$ directly maps the interaction history to an executable action and may overlook explicit reasoning about latent states and environment dynamics. Recent VLM agent frameworks therefore incorporate prospective world modeling into the reasoning process, allowing the policy model to internally simulate the consequence of a candidate action before interacting with the environment~\cite{wang2025vagen,xing2025critiquesworldmodels}. Specifically, at turn $t$, the agent first summarizes the partial observation $o_t$ and history $h_t$ into a textual belief state $\hat{s}_t$, which serves as an internal approximation of the underlying latent state. 
Conditioned on $\hat{s}_t$ and the task goal $g$, the agent proposes a candidate action $a^{\prime}_t$ for mental rehearsal and predicts the corresponding future belief state, i.e., 
$s^{\prime}_{t+1} \sim \pi_\theta(\cdot|\hat{s}_t, a^{\prime}_t)$.
Here, $s^{\prime}_{t+1}$\footnote{For simplicity, at the turn of $t + 1$, we re-generate state belief $\hat{s}_{t+1}$ given the parallel observation $o_{t+1}$, rather than using the predicted state belief in the $t$-th turn.} denotes the imagined next state predicted before execution.
This prospective simulation forms an internal reasoning trace:
\begin{equation}
\small
\texttt{<Obs>}\hat{s}_t\texttt{</Obs>} \texttt{<Res>}a^{\prime}_t\texttt{</Res>} \texttt{<Pred>}s^{\prime}_{t+1}\texttt{</Pred>}.
\nonumber
\end{equation}

The final executable action $a_t$ ($\texttt{<Ans>}a_t\texttt{</Ans>}$) is then sampled by conditioning the policy on both the interaction history and the above prospective reasoning trace.

\section{Method}\label{sec:method}
\begin{figure*}[t]
\centering
\includegraphics[width = 1.0\textwidth]{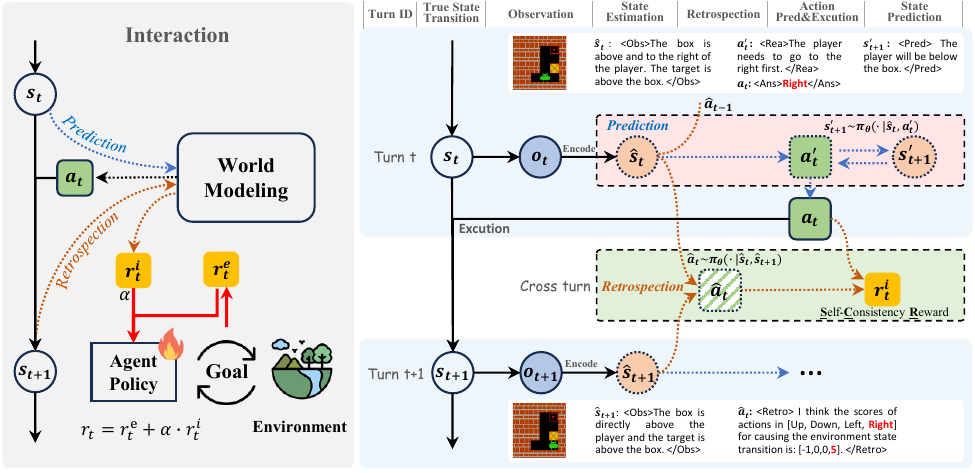}
\caption{
Overview of the RWM framework. At turn $t$, the agent first infers the current belief state $\hat{s}_t$ from the observation and performs retrospection over the previous transition $(\hat{s}_{t-1}, \hat{s}_t)$ and \hl{get attributed action $\hat{a}_{t-1}$}. It then conducts prospective reasoning to propose a candidate action $a'_t$ and predict an imagined next state $s'_{t+1}$ before producing the executable action $a_t$. 
\hl{After observing the next belief state $\hat{s}_{t+1}$, RWM performs a cross-turn attribution step, inferring $\hat{a}_{t}$ from the realized transition $(\hat{s}_t, \hat{s}_{t+1})$, and evaluates its consistency with the executed action $a_t$.} The resulting self-consistency reward is then assigned back to turn $t$ for policy optimization.
}
\label{fig:framework}
\end{figure*}

Prospective world modeling enables VLM agents to imagine future states before execution, but it does not directly verify whether an executed action is consistent with the observed state transition. For each transition from turn $t$ to turn $t+1$, after the agent executes $a_t$ and receives the next observation, the resulting belief transition $(\hat{s}_t, \hat{s}_{t+1})$ provides a natural opportunity to evaluate whether $a_t$ is a plausible cause of the observed change. Equivalently, at the beginning of turn $t+1$, this verification appears as a retrospective attribution step over the previous transition. Motivated by this observation, we propose Retrospective World Modeling (RWM), a new agent learning framework that uses retrospective action attribution as a consistency verification mechanism for policy learning. As shown in Fig.~\ref{fig:framework}, RWM first performs retrospective attribution over the observed transition, then converts the attribution result into a margin-based intrinsic reward, and finally optimizes the policy with temporally aligned RL.
\subsection{Retrospective World Modeling}
RWM extends the prospective reasoning loop with a retrospective verification step. At the beginning of turn $t$, the agent receives a new observation $o_t$ and summarizes the interaction context into a current belief state $\hat{s}_t$. Since the belief state from the previous turn, $\hat{s}_{t-1}$, is already available, the agent can reason backward over the observed transition $(\hat{s}_{t-1}, \hat{s}_t)$ and infer which action could have caused this change. Formally, the retrospectively attributed action is sampled as:
\begin{equation}
    \hat{a}_{t-1} \sim \pi_\theta(\cdot \mid \hat{s}_{t-1}, \hat{s}_t).
    \label{eq:retro_attribution}
\end{equation}
Here, $\hat{a}_{t-1}$ is a model-internal explanation of the previous transition rather than a ground-truth label. It is used to verify the consistency between the observed transition and the action actually executed in the previous turn. To avoid trivial copying from the interaction history, retrospective attribution is conditioned only on the two belief states involved in the transition, rather than on the full action history. For the first turn, where no previous transition is available, the retrospective field is filled with a \texttt{null} token and is excluded from reward computation.

After the retrospective step, the agent proceeds to prospective reasoning for the current turn. Conditioned on the current belief state $\hat{s}_t$, the task goal $g$, and the retrospective context $\hat{a}_{t-1}$, the policy proposes a candidate action $a^{\prime}_t$ for mental rehearsal and predicts the imagined next belief state $s^{\prime}_{t+1}$.
Together, the insights from both the retrospective grounding and prospective simulation yield a new trajectory before the agent outputs the final action $a_t$ for the environment to execute:
\begin{equation}
\small
\texttt{<Obs>}\hat{s}_t\texttt{</Obs>} 
\texttt{<Retro>}\hat{a}_{t-1}\texttt{</Retro>} 
\texttt{<Res>}a^{\prime}_t\texttt{</Res>} \texttt{<Pred>}s^{\prime}_{t+1}\texttt{</Pred>}.
\label{eq:trajectory}
\end{equation}
In this trace, the \texttt{<Retro>} field explains the transition that has already occurred, while the \texttt{<Rea>} and \texttt{<Pred>} fields support planning for the current action. 
This ``Observe-Retrospect-Predict-Act'' loop makes retrospective verification an explicit part of multi-turn decision-making, while preserving prospective simulation for planning.

\subsection{Self-Consistency as Intrinsic Reward}
\label{sec:method_reward_formulation}
The retrospective attribution in Eq.~\ref{eq:retro_attribution} provides a qualitative explanation of the previous transition, but using the sampled action token directly as a reward signal is unreliable. A direct token match between the sampled attribution and the executed action gives a sparse signal, and it is also vulnerable to shortcut learning when the previous action appears in the interaction history. To obtain a denser and more robust learning signal, we formulate retrospective verification as a discriminative scoring problem over the finite action space $\mathcal{A}$.

\textbf{Discriminative Action Scoring}. After the transition from turn $t$ to turn $t+1$ is observed, RWM evaluates the realized transition $(\hat{s}_t, \hat{s}_{t+1})$ with a history-agnostic scoring interface implemented by the same VLM policy. Specifically, the model scores each candidate action according to how plausible it is as the cause of the observed transition:
\begin{equation}
\mathbf{v}_t = F_\theta(\hat{s}_t, \hat{s}_{t+1}; \mathcal{A}) \in \mathbb{R}^{|\mathcal{A}|},
\label{eq:score_vector}
\end{equation}
where $F_\theta$ denotes the retrospective scoring form of the policy model, and each entry $\mathbf{v}_t^{(a)}$ measures the plausibility of action $a \in \mathcal{A}$ as the cause of the transition from $\hat{s}_t$ to $\hat{s}_{t+1}$. In practice, these scores are derived from the logits assigned to candidate action tokens under a prompt that exposes only the state transition and the candidate action set.

\textbf{Self-Consistency Reward (SCR)}. 
Given the score vector, we define the self-consistency reward as the normalized margin between the score of the executed action $a_t$ and the strongest alternative action:
\begin{equation}
r_t^i =
\frac{
\mathbf{v}_t^{(a_t)} -
\max_{a \in \mathcal{A} \setminus \{a_t\}} \mathbf{v}_t^{(a)}
}{
S_{\max} - S_{\min}
}.
\label{eq:intrinsic_reward}
\end{equation}
Here, $S_{\max}$ and $S_{\min}$ denote the upper and lower bounds of the scoring scale and are used to normalize the reward magnitude. A positive margin indicates that the executed action is scored as the most plausible explanation of the observed transition, while a negative margin indicates that another candidate action better explains the transition. Compared with direct action reconstruction, this margin-based formulation provides a dense signal and explicitly contrasts the executed action with the hardest distractor, encouraging sharper action-transition consistency.

\textbf{Total Reward}. The total reward for policy optimization is defined as a composite sum:
\begin{equation}
r_t = r_t^e + \alpha \cdot r_t^i,
\label{eq:total_reward}
\end{equation}
where $r_t^e$ denotes the extrinsic reward, including task success, format compliance, and step penalties, and $\alpha$ controls the strength of retrospective consistency regularization. In Eq.~\ref{eq:total_reward}, the extrinsic reward provides task-level supervision, while \texttt{SCR} provides transition-level feedback that regularizes the policy toward actions whose effects can be consistently explained by the observed state change.

\subsection{Policy Optimization}
\label{sec:policy_optimization}
We optimize the policy $\pi_\theta$ using Proximal Policy Optimization (PPO~\cite{schulman2017proximalpolicyoptimizationalgorithms}). During training, the critic $V_\phi$ is implemented as a scalar value head on the final hidden state of the VLM. Since each trajectory contains both environmental prompts and model-generated tokens, we apply a binary mask $\mathcal{M}$ so that policy gradients are computed only on generated reasoning and action tokens, while environmental prompts are excluded from the policy update.

\textbf{Temporal Reward Assignment}. 
The self-consistency reward is delayed by one turn because it requires the next belief state $\hat{s}_{t+1}$ to be observed. Therefore, although the score vector in Eq.~\ref{eq:score_vector} is computed at the beginning of turn $t+1$, the resulting reward $r_t^i$ is assigned back to the action $a_t$ executed at turn $t$: $r_t^i \leftarrow \mathrm{SCR}(\hat{s}_t, \hat{s}_{t+1}, a_t)$. 
This temporal assignment aligns the retrospective verification signal with the transition that generated it and preserves the standard turn-level reward structure of the underlying POMDP.

\textbf{Dense Reward Integration and Configurations.} Given the structured reasoning trajectory established earlier, training the agent with standard PPO relies solely on sparse task and format rewards. In this standard setup, credit assignment is handled by the conventional Generalized Advantage Estimation (GAE)~\cite{schulman2015high}, which propagates sparse rewards backward across the established trajectory (Eq.~\ref{eq:trajectory}) without explicit turn boundaries. We define this sparse-reward configuration as \textbf{RWM-Base} setting.

However, multi-turn VLM agents face a fundamental granularity gap: our composite reward $r_t$ is defined at the macro-level of environmental turns, whereas PPO requires micro-level token advantages. To seamlessly backpropagate turn-wise semantic feedback, we go a step further. Following VAGEN~\cite{wang2025vagen}, we adopt a Bi-Level GAE protocol (detailed formulations are provided in Appendix.~\ref{sec:append_advantage_estimation}). This mechanism bridges the gap by first computing turn-level advantages and then injecting them into the micro-scale token generation, allowing for dense reward assignment.
By employing this Bi-Level GAE for turn-aware credit assignment, we can effectively inject dense supervision into the reasoning steps. Alongside external LLM-as-a-Judge~\cite{wang2025vagen} supervision for visual state evaluation, we critically integrate our intrinsic Self-Consistency Reward ($r_t^i$) to explicitly penalize causal hallucinations at each turn. This comprehensive, dense-reward configuration establishes our \textbf{RWM-Full} setting.

\textbf{Optimization Objective.} Let $\tau$ denote the trajectory and $\mathcal{M}$ be the mask that isolates generated action tokens (excluding environmental prompts). The above process yields the specific token advantage $A_i$ (Eq.~\ref{eq:token_advantage}) used in the PPO clipping objective $\mathcal{L}(\theta)$:
 
\begin{equation}
\label{eq:PPO}
    \mathbb{E}_{(\tau) \sim \pi_{\text{old}}} \left[ \frac{\sum_{i} \mathcal{M}_i \cdot \min \left( \rho_i(\theta) A_i, \text{clip}(\rho_i(\theta), 1 - \epsilon, 1 + \epsilon) A_i \right)}{\sum_k \mathcal{M}_k}  \right]
\end{equation}
where $\rho_i(\theta) = \frac{\pi_\theta(\tau_i | \tau_{<i})}{\pi_{\text{old}}(\tau_i | \tau_{<i})}$ is the probability ratio and $\epsilon$ is the clipping parameter. This optimization ensures the agent's policy is updated stably, driven by both the intrinsic consistency verification and extrinsic task goals.

\section{Experiments}
\subsection{Experimental Settings}
\label{sec:experiment_setting}
\textbf{Environments.}
To analyze the visual reasoning capabilities of VLM agents, we evaluate on four distinct agentic tasks. These benchmarks cover challenges including diverse visual state representations and action spaces: 2D grid puzzles (\textit{Sokoban}~\cite{SchraderSokoban2018} and \textit{FrozenLake}~\cite{towers2024gymnasium}), embodied 3D navigation (\textit{Navigation}~\cite{ai2thor}) and embodied object manipulation (\textit{ManiSkill}~\cite{tao2025maniskill,hiranaka2023primitive}).
Sokoban requires pushing boxes to targets while avoiding deadlocks, and FrozenLake involves navigating to a goal under hazardous conditions (with deterministic dynamics).
Navigation is a first-person indoor search task that demands spatial reasoning from partial observations, while ManiSkill focuses on robotic manipulation with a Panda arm using a hybrid action space (e.g., \texttt{pick(x,y,z)}), requiring precise visual grounding from 3D scenes to actionable coordinates.
For Navigation, we assess performance on both \textit{Base} and \textit{Common Sense} evaluation sets.
For Sokoban, we employ the standard 6 $\times$ 6 grid and further expand to a \textit{Hard} setting (scaled to large dimensions) to probe generalization under increased complexity.
More details are provided in Appendix.~\ref{sec:append_setting_env}.

\begin{table}[!t] 
    \centering
    \footnotesize
    \caption{Main results on the 4 general agentic benchmarks. We report average task success rates across Sokoban, Navigation, ManiSkill, and FrozenLake benchmarks. We adopt Qwen2.5-VL-3B as the backbone for the trained models. The best and second-best performance among open-source models/methods are highlighted in \textbf{bold} and \underline{underline}, respectively.}
    \begin{adjustbox}{width=\linewidth}
    \setlength\tabcolsep{1pt}
    \setlength\extrarowheight{3pt}
    \arrayrulecolor[gray]{0.7} 
    \begin{tabular}{l|cc|c|cc|c|cccc|c|c|c}
        \toprule
        \multirow{2}{*}{\textbf{Models}}
        & \multicolumn{3}{c|}{\textbf{Sokoban}}
        & \multicolumn{3}{c|}{\textbf{Navigation}} 
        & \multicolumn{5}{c|}{\textbf{ManiSkill}} 
        & \multirow{2}{*}{\textbf{Frozen}} 
        & \multirow{2}{*}{\textbf{Overall}} \\ 
        
        \cmidrule(lr){2-4} \cmidrule(lr){5-7} \cmidrule(lr){8-12}
        &Standard &Hard & \textbf{Average} & Base & Common & \textbf{Average} & Place & Stack & Drawer & Align & \textbf{Average} & &  \\ 
        \hline
        \multicolumn{14}{c}{Open-Source Models} \\
        \hline
        Qwen2.5-VL-72B~\cite{bai2023qwenvlversatilevisionlanguagemodel}
        &0.18 &- &0.18
        &\textbf{0.72} &\textbf{0.75} &\textbf{0.74} 
        &\textbf{1.00} &0.50 &0.00 &\textbf{1.00} &0.63  
        &0.44  
        &0.51  \\

        Qwen2.5-VL-7B~\cite{bai2023qwenvlversatilevisionlanguagemodel}
        &0.13 &- &0.13
        &0.28 &0.39 &0.34 
        &0.00 &0.00 &0.00 &\underline{0.75} &0.19  
        &0.14  
        &0.27  \\

        VLM-R1-3B~\cite{shen2025vlm} 
        &0.13 &- &0.13
        &0.31 &0.34 &0.33 
        &0.00 &0.00 &0.00 &0.00 &0.00  
        &0.13  
        &0.23  \\
        
        \hline
        \rowcolor{gray!25} \multicolumn{14}{c}{Beyond Prediction: Steering VLM Agents with Retrospective World Modeling (Backbone: Qwen2.5-VL-3B)} \\
        \hline
        
        \rowcolor{gray!8}
        Qwen2.5-VL-3B~\cite{bai2023qwenvlversatilevisionlanguagemodel}
        &0.06 &0.05 &0.06
        &0.22 &0.27 &0.25 
        &0.00 &0.00 &0.00 &0.00 &0.00  
        &0.09  
        &0.08  \\
        
        \rowcolor{gray!8}
        + Vanilla-PPO~\cite{ragen} 
        &0.18 &0.13 &0.16
        &0.32 &0.25 &0.29 
        &0.00 &0.00 &0.00 &0.00 &0.00  
        &0.21  
        &0.12  \\

        \rowcolor{gray!8}
        + ReAct-RL~\cite{yao2023react} 
        &0.27 &0.20 &0.24
        &0.39 &0.37 &0.38 
        &0.00 &0.00 &0.00 &0.00 &0.00  
        &0.30  
        &0.17  \\

        \rowcolor{gray!8}
        + VAGEN-Base~\cite{wang2025vagen} 
        &0.41 &0.33 &0.37
        &0.47 &0.51 &0.49 
        &\underline{0.88} &0.63 &\underline{0.63} &\underline{0.75} &0.72  
        &0.43  
        &0.56  \\

        \rowcolor{gray!8}
        \textbf{+ RWM-Base (Ours)}  
        &\underline{0.58} &\underline{0.46} &\underline{0.52}
        &\underline{0.67} &0.64 &0.66
        &\textbf{1.00} &\underline{0.88} &\textbf{0.88} &\textbf{1.00} &\underline{0.94} 
        &\underline{0.75}  
        &\underline{0.76}  \\

        \hline
        \rowcolor{gray!25} \multicolumn{14}{c}{World Modeling with Bi-Level GAE and LLM-as-a-Judge (Using GPT-4.1 nano)} \\
        \hline
        \rowcolor{gray!8}
        + VAGEN-Full 
        &0.52 &\underline{0.46} &0.48
        &0.54 &0.56 &0.55 
        &\textbf{1.00} &\underline{0.88} &\textbf{0.88} &\textbf{1.00} &\underline{0.94}  
        &0.54  
        &0.71  \\
        
        \rowcolor{gray!8}
        \textbf{+ RWM-Full (Ours)}  
        &\textbf{0.69} &\textbf{0.63} &\textbf{0.66}
        &\underline{0.67} &\underline{0.69} &\underline{0.68} 
        &\textbf{1.00} &\textbf{1.00} &\textbf{0.88} &\textbf{1.00} &\textbf{0.97}  
        &\textbf{0.77}  
        &\textbf{0.81}  \\

        \hline
        \rowcolor{gray!25} \multicolumn{14}{c}{\gr Proprietary Models} \\
        \hline

        \gr GPT-4o~\cite{openai_gpt4o_systemcard_2024}  
        &\gr0.43 &\gr0.38 &\gr0.41
        &\gr0.75 &\gr0.69 &\gr0.72 
        &\gr0.50 &\gr0.63 &\gr0.00 &\gr0.88 &\gr0.50
        &\gr0.54 
        &\gr0.53 \\

        \gr GPT-5~\cite{openai_gpt5_2025} 
        &\gr0.70 &\gr0.62 &\gr0.66
        &\gr0.75 &\gr0.81 &\gr0.78 
        &\gr1.00 &\gr0.63 &\gr0.00 &\gr1.00 &\gr0.66
        &\gr0.77 
        &\gr0.70 \\

        \gr o4-mini~\cite{openai_gpt_o3_o4_mini_2025}
        &\gr0.44 &\gr0.40 &\gr0.42
        &\gr0.75 &\gr0.75 &\gr0.75 
        &\gr1.00 &\gr0.50 &\gr0.00 &\gr0.75 &\gr0.56
        &\gr0.82 
        &\gr0.60 \\
        
        \gr Claude 4.5 Sonnet~\cite{anthropic_claude45_2025}  
        &\gr0.31 &\gr0.26 &\gr0.29
        &\gr0.67 &\gr0.67 &\gr0.67 
        &\gr0.63 &\gr0.50 &\gr0.00 &\gr1.00 &\gr0.53
        &\gr0.80 
        &\gr0.54 \\

        \gr Claude 3.7 Sonnet~\cite{TheC3}
        &\gr0.25 &\gr0.18 &\gr0.22
        &\gr0.48 &\gr0.47 &\gr0.48 
        &\gr0.63 &\gr0.13 &\gr0.00 &\gr1.00 &\gr0.44
        &\gr0.69 
        &\gr0.43 \\

        \gr Gemini 2.5 Pro~\cite{google_gemini25_2025}  
        &\gr0.58 &\gr0.42 &\gr0.50
        &\gr0.63 &\gr0.63 &\gr0.63 
        &\gr0.63 &\gr0.63 &\gr0.00 &\gr0.75 &\gr0.50
        &\gr0.78 
        &\gr0.56 \\
        
        \bottomrule
    \end{tabular}
    \end{adjustbox}
    \label{tab:main_results}
\end{table}

\textbf{Baselines.} 
We adopt Qwen2.5-VL-3B~\cite{bai2023qwenvlversatilevisionlanguagemodel} as the VLM backbone.
We compare our approach against two categories of baselines.
(1) Open-source methods:
\textbf{Vanilla-PPO}~\cite{ragen} directly fine-tunes the base model using PPO; 
\textbf{ReAct-RL} follows the ReAct~\cite{yao2023react} paradigm of interleaved reasoning traces (``observe-reason-act'') optimized via PPO; 
\textbf{VAGEN-Base}~\cite{wang2025vagen} structures reasoning into explicit \textit{prospective} world modeling (state estimation and future transition prediction) but relies on standard token-level GAE and sparse task rewards; 
\textbf{VAGEN-Full}~\cite{wang2025vagen} enhances this prospective paradigm by employing Bi-Level GAE for credit assignment and dense reasoning supervision provided by an external LLM-as-a-Judge. 
(2) Proprietary Models: We also report zero-shot results from state-of-the-art proprietary models, including \texttt{GPT-5}~\cite{openai_gpt5_2025}, \texttt{GPT-4o}~\cite{openai_gpt4o_systemcard_2024}, \texttt{o4-mini}~\cite{openai_gpt_o3_o4_mini_2025}, \texttt{Claude 4.5 Sonnet}~\cite{anthropic_claude45_2025}, \texttt{Claude 3.7 Sonnet}~\cite{TheC3}, and \texttt{Gemini 2.5 Pro}~\cite{google_gemini25_2025}. 
More details are provided in Appendix.~\ref{sec:append_setting_baselines}.

\textbf{Evaluation Metrics.} 
Success rate is the primary metric. To evaluate how effectively our retrospective constraints filter out causally implausible hallucinations, we additionally report action validness (actions within the defined space) and action effectiveness (actions inducing meaningful state changes toward the goal, e.g., avoiding blocking walls), alongside standard rewards.

\textbf{Implementation Details.}
We use \texttt{Qwen2.5-VL-3B}~\cite{bai2023qwenvlversatilevisionlanguagemodel} as the backbone. Global training batch size is set to 128, with a learning rate of $1 \times 10^{-6}$ for the actor and $1 \times 10^{-5}$ for the critic.
For evaluation, we set the generation temperature to 1.0.
Results are averaged over 3 runs on 128 diverse test cases to ensure statistical significance.
All experiments were conducted on 4$\times$ NVIDIA H800 (80GB) GPUs.

\subsection{Main Results}
\textbf{Main Performance.} 
Tab.~\ref{tab:main_results} reports success rates across four diverse environments, revealing a consistent performance hierarchy. RWM achieves substantial overall gains over model-free approaches (0.76 vs. 0.12 for Vanilla-PPO), and significantly outperforms ReAct-RL (0.17), confirming that explicit world modeling provides superior guidance over interleaved reasoning traces alone. Crucially, we compare against the prospective-only VAGEN across two configurations. In the Base setting, RWM-Base demonstrates a remarkable advantage, boosting the overall average success rate from 0.56 (VAGEN-Base) to 0.76. This superiority persists even with dense supervision: RWM-Full lifts the performance from 0.71 (VAGEN-Full) to an impressive 0.81. This sharp contrast suggests that our retrospective constraint effectively filters out causally implausible transitions often hallucinated by purely prospective models. Furthermore, despite utilizing a significantly smaller backbone (\texttt{Qwen2.5-VL-3B}), RWM surpasses several powerful proprietary models like GPT-4o (0.53) and Claude 4.5 Sonnet (0.54), and even outperforms GPT-5 (0.70), highlighting the potential of specialized agents to unlock robust reasoning capabilities through self-supervised grounding.

\textbf{Generalization on Challenging Tasks.} 
To probe robustness, we evaluate agents trained exclusively on standard maps on the unseen \textit{Sokoban-Hard} setting. While baselines suffer severe degradation under this distribution shift, RWM exhibits superior generalization with a $0.46$ success rate, significantly outperforming VAGEN ($0.33$). This resilience confirms that retrospective attribution forces the agent to internalize scalable physical rules rather than merely memorizing training layouts.

\subsection{Ablation Study}
\label{sec:ablation}
In this section, we conduct extensive ablations to dissect our framework, aiming to validate retrospective signal internalization, justify our reward design, and show how to prevent reward hacking.

\begin{wrapfigure}{r}{0.55\linewidth}
  \centering
  \vspace{-0.3cm}
  \begin{subfigure}{0.49\linewidth}
    \centering
    \includegraphics[width=\linewidth]{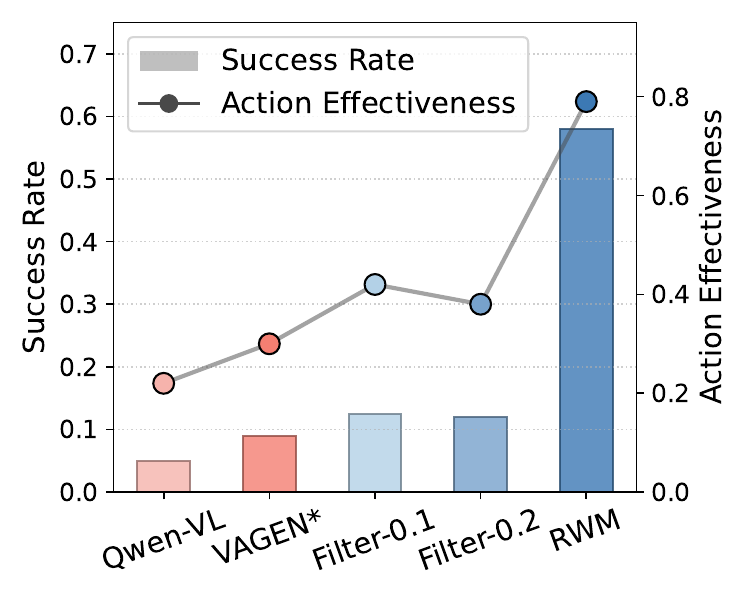}
    \caption{Sokoban}
  \end{subfigure}
  \begin{subfigure}{0.49\linewidth}
    \centering
    \includegraphics[width=\linewidth]{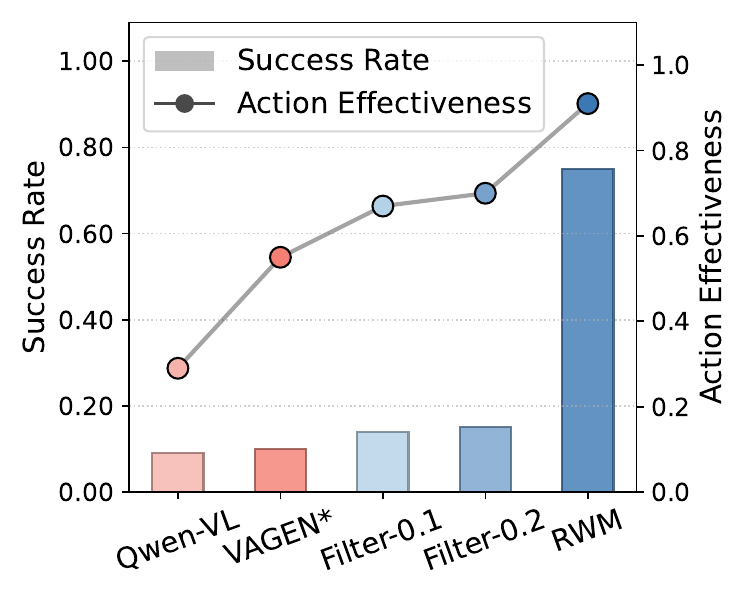}
    \caption{FrozenLake}
  \end{subfigure}
  \caption{
   Ablation on signal internalization. RWM-Filter applies the intrinsic reward $r^i_t$ purely as a training-free rejection sampling threshold ($\delta=0.1, 0.2$) to filter inconsistent actions. Results show that explicit RL internalization yields substantially higher success rates and action effectiveness. (*: val before train)
  }
  \label{fig:wm_filter}
\end{wrapfigure}

\textbf{Is the RL-based ``Internalization'' Necessary?}  
A core premise of our method is that the retrospective signal should be internalized into policy parameters. 
We investigate this via \textbf{RWM-Filter}, a training-free baseline using our intrinsic reward as a hard constraint for rejection sampling during inference. Fig.~\ref{fig:wm_filter} reveals two critical insights. First, RWM-Filter consistently outperforms the vanilla VAGEN (e.g., +5\% success rate), validating our consistency signal as a reliable indicator of physical grounding. Second, the RL-internalized RWM drastically outperforms RWM-Filter (over 40\% gains). This confirms that while post-hoc filtering helps, embedding this causal constraint directly into policy parameters via RL is crucial to improve policy exploitation, transforming ``\textit{checking for errors}'' into ``\textit{learning not to make them}''.

\begin{wrapfigure}{r}{0.3\linewidth}
  \centering
  \vspace{-0.3cm}
  \includegraphics[width=\linewidth]{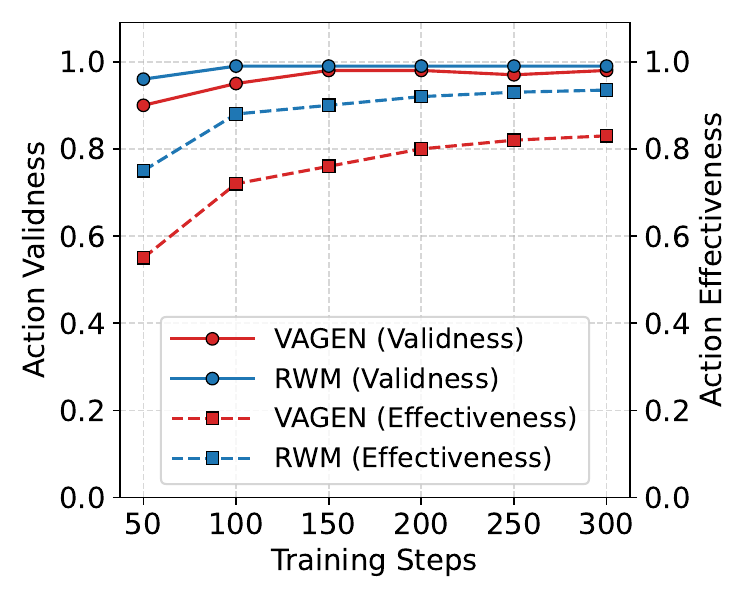}
  \caption{Ablation on action validness and effectiveness.}
  \vspace{-15pt}
  \label{fig:ablation_action_validness_and_effectiveness}
\end{wrapfigure}

\textbf{Mitigating Causal Hallucinations.}
To validate our core motivation that retrospective verification reduces hallucinations, we compare the training dynamics of action validness and effectiveness on Sokoban (Fig.~\ref{fig:ablation_action_validness_and_effectiveness}). While the prospective-only baseline (VAGEN) gradually learns valid action formats, it severely struggles with effectiveness, often executing physically meaningless moves like pushing against blocking walls (peaking at only 0.8). In contrast, RWM rapidly achieves near-perfect validness and consistently maintains superior effectiveness (>0.95). This massive margin directly demonstrates that our retrospective constraint provides robust physical grounding, explicitly steering the agent away from causally implausible hallucinations toward purposeful and executable actions.
\begin{figure}[t]
  \centering
  \begin{minipage}{0.48\linewidth}
    \centering
    \begin{subfigure}{0.49\linewidth}
      \includegraphics[width=\linewidth]{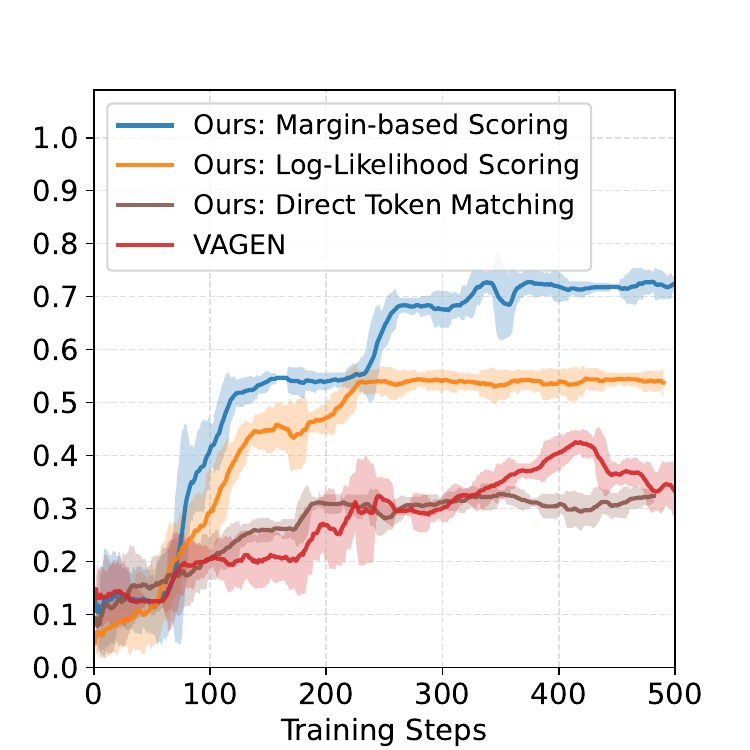}
      \caption{\footnotesize Success Rate}
    \end{subfigure}
    \begin{subfigure}{0.49\linewidth}
      \includegraphics[width=\linewidth]{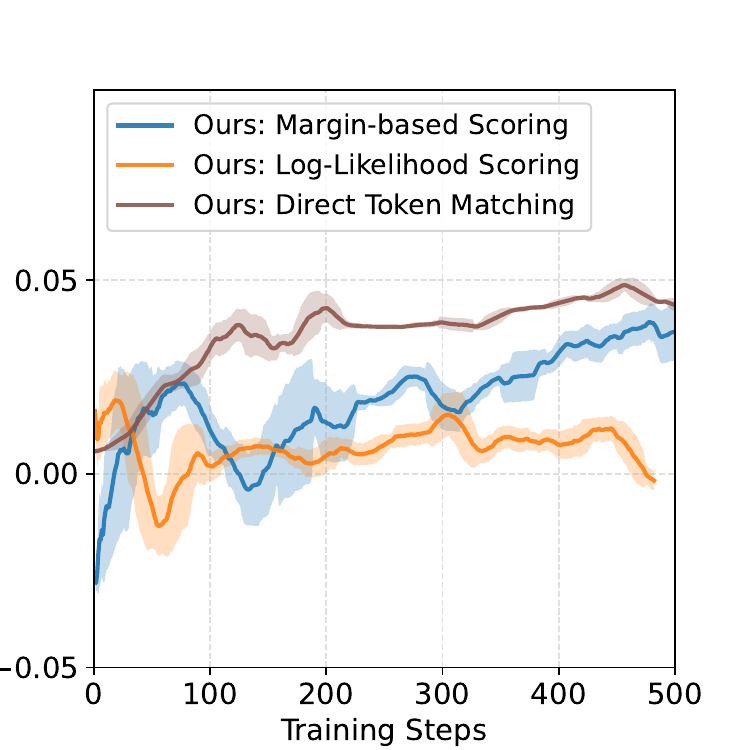}
      \caption{\footnotesize Self-Consistency Reward}
    \end{subfigure}

    \captionof{figure}{Ablation on reward design on FrozenLake benchmark. We show the training success rate and intrinsic reward curves, comparing RWM with three reward designs and VAGEN.}
    \label{fig:ablation_reward_design}
  \end{minipage}
  \hfill
  \begin{minipage}{0.48\linewidth}
    \centering
    \begin{subfigure}{0.49\linewidth}
      \includegraphics[width=\linewidth]{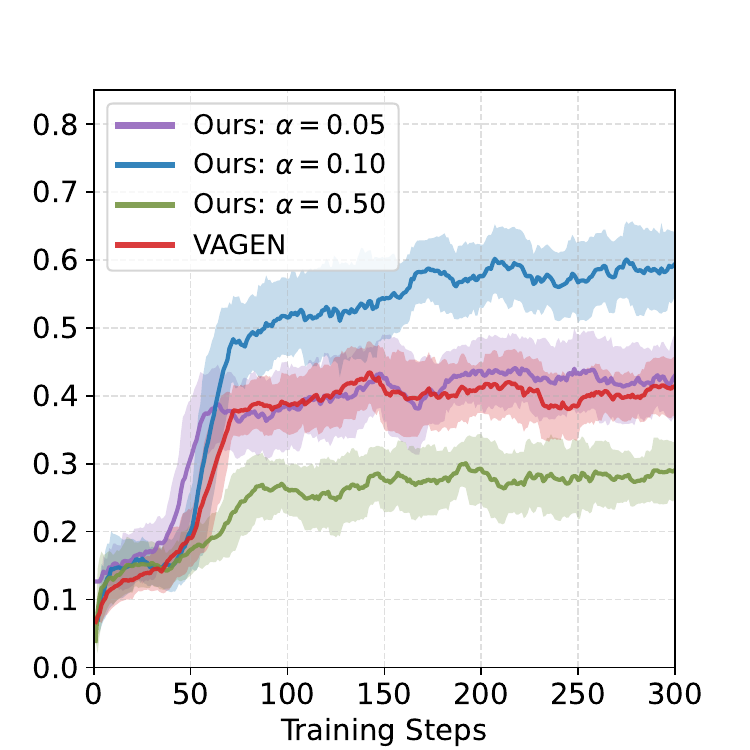}
      \caption{\footnotesize Success Rate}
    \end{subfigure}
    \begin{subfigure}{0.49\linewidth}
      \includegraphics[width=\linewidth]{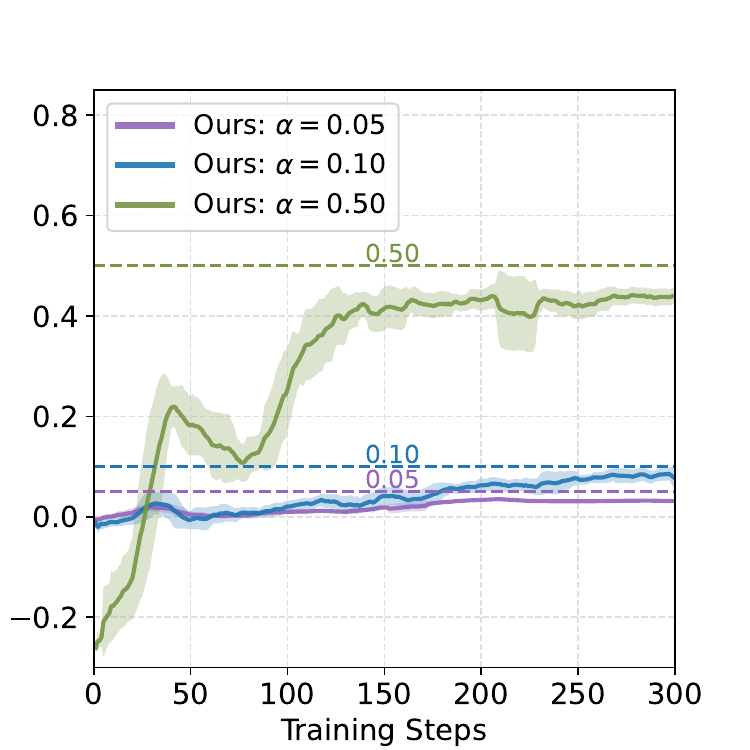}
      \caption{\footnotesize Self-Consistency Reward}
    \end{subfigure}

    \captionof{figure}{Ablation on reward weight $\alpha$ on Sokoban benchmark. We show the training success rate and intrinsic reward curves. A large $\alpha$ tends to overfit, while a small $\alpha$ limits gains.}
    \label{fig:ablation_reward_weight}
  \end{minipage}
\end{figure}

\textbf{Intrinsic Reward Design.}
To balance reward density with robustness against shortcut learning, we compare against two alternatives in Fig.~\ref{fig:ablation_reward_design}. 
\textbf{1) Direct Token Matching} assigns a binary reward if the sampled token $\hat{a}_{t}$ matches $a_{t}$. It yields poor success rates ($\approx$0.30) and the reward saturate quickly (near the 0.05 upper bound), as sparse binary feedback hinders optimization and encourages shortcut learning, simply copying action history.
\textbf{2) Log-Likelihood Scoring} (normalized log-probability of the ground-truth action) improves performance ($\approx$0.53) but suffers from high variance, as massive penalties for low-probability events destabilize training.
\textbf{3) Margin-based Scoring} (Ours) avoids these pitfalls by penalizing the gap between the ground-truth action and the hardest distractor (Eq.~\ref{eq:intrinsic_reward}). This provides a dense, bounded, and stable signal that effectively curbs shortcut learning and maximizes causal confidence, yielding the fastest convergence and highest success rate.
More details are in Appendix.~\ref{sec:append_reward_design}. 

\textbf{Sensitivity to Retrospection Weight.} 
The parameter $\alpha$ (Eq.~\ref{eq:total_reward}) balances retrospective grounding with prospective planning. As Fig.~\ref{fig:ablation_reward_weight} shows, a minimal weight ($\alpha=0.05$) provides insufficient gradients, yielding negligible gains over VAGEN. Conversely, an excessive weight ($\alpha=0.50$) halves the success rate (from $0.58$ to $0.29$). In this regime, the intrinsic reward saturates early (Fig.~\ref{fig:ablation_reward_weight}(b)), indicating a form of reward hacking where the agent overfits to explaining transitions at the expense of task completion. This confirms the retrospective signal functions best as a regularizer. 

\begin{wraptable}{r}{0.45\linewidth}
\centering
\vspace{-0.40cm}
\caption{
Ablation on history leakage under 3 context regulation strategies. 
}
\label{tab:leakage}
\begin{adjustbox}{width=\linewidth}
\setlength\tabcolsep{3pt}
\setlength\extrarowheight{2pt}
\begin{tabular}{l|c|c}
\toprule
\textbf{Method} & \textbf{Sokoban} & \textbf{FrozenLake} \\
\midrule
Qwen2.5-VL-3B & 0.06 & 0.09 \\
VAGEN         & 0.41 & 0.43 \\
\midrule
\textbf{RWM} & & \\
\quad \textit{- w/ Regulation-free}        & 0.39 & 0.40 \\
\quad \textit{- w/ Dropout ($\beta=0.10$)} & 0.44 & 0.52 \\
\quad \textit{- w/ Dropout ($\beta=0.20$)} & 0.50 & 0.59 \\
\quad \textit{- w/ Dropout ($\beta=0.50$)} & 0.33 & 0.38 \\
\quad \textbf{\textit{- History-agnostic}} 
& \textbf{0.58} & \textbf{0.75} \\
\bottomrule
\end{tabular}
\end{adjustbox}
\end{wraptable}

\textbf{Historical Context Leakage.} 
In principle, RWM requires that the agent deduce $\hat{a}_{t}$ solely from the state transition $(\hat{s}_{t}, \hat{s}_{t+1})$. However, in multi-turn RL, full history access risks historical context leakage: the agent trivially retrieves the previous action $a_{t}$ from memory rather than inferring it via world dynamics, essentially reward hacking. To enforce genuine causal reasoning, we evaluate three regulation strategies in Tab.~\ref{tab:leakage}.
\textbf{1) Regulation-free} allows unrestricted history access. It performs comparably to VAGEN, confirming that memory shortcuts render the retrospective signal ineffective. 
\textbf{2) Action Dropout} randomly occludes action-related tokens in the history with probability $\beta$ (e.g., replacing them as \texttt{...<Rea>[MASKED]</Rea>...<Ans>[MASKED]</Ans>}). While mild dropout ($\beta=0.1, 0.2$) yields modest gains, aggressive masking ($\beta=0.5$) severely disrupts the semantic coherence required for long-horizon planning. 
\textbf{3) History-agnostic Inference (Ours)} strictly isolates the retrospective prompt, conditioning deduction solely on the observed state change. Synergizing with our Margin-based Scoring, which resists simple token copying by requiring evaluation over the full action space, this strict information bottleneck completely cuts off memory exploitation, achieving the highest success rate. We adopt this setting for all RWM experiments.

\section{Conclusion}
In this paper, we challenge the conventional prospective-only world modeling, arguing that a robust agent requires not only predicting future outcomes but also serving as a retrospective analyst. 
To this end, we propose Retrospective World Modeling (RWM), a novel cross-turn framework that equips VLM agents with a bidirectional understanding of environmental dynamics and the capability of deducing underlying causality behind the observed transitions. By formulating the Self-Consistency Reward (SCR), we convert the retrospective causal attribution into a dense intrinsic signal, effectively steering the agent towards physically grounded and logically consistent behaviors.
Our extensive experiments and ablation studies demonstrate the effectiveness of our method and the necessity of internalizing
this intrinsic signal into policy. Furthermore, we assess various reward formulations and context regulation strategies to prevent history leakage, ensuring the agent learns genuine physical rules rather than exploiting memory shortcuts.
Future work will extend this paradigm to address the challenge of causal ambiguity over longer horizons.

\bibliographystyle{unsrt}
\bibliography{main}

\newpage
\section*{\huge Appendix}
\appendix

\section{Details of Reward Design}
\label{sec:append_reward_design}

In the main text (Section \ref{sec:method_reward_formulation}), we introduced Margin-based Scoring as our primary method for quantifying the self-consistency intrinsic reward. To validate the superiority of this design, our ablation study (Section \ref{sec:ablation}) compares it against two alternative formulations: Direct Token Matching and Log-Likelihood Scoring. In this section, we provide the formal definitions and detailed implementations for these alternative designs.

\textbf{Relationships Among Reward Designs.} It should be noted that both Log-Likelihood Scoring and Margin-based Scoring are discriminative action scoring methods, meaning they use the same scoring mechanism. The difference lies in how the resulting score vector is converted into a reward. In contrast, Direct Token Matching does not require any scoring.
\subsection{Direct Token Matching}

In this variant, we simplify the retrospective process by enforcing a discrete decision rather than evaluating the full probability distribution. Instead of calculating a margin-based score over the entire action space, the agent is prompted to directly sample a single action token that explicitly identifies the cause of the transition.

Formally, given the observed transition tuple $(\hat{s}_t, \hat{s}_{t+1})$, the policy samples a predicted retrospective action $\hat{a}_{t}$:
\begin{equation}
    \hat{a}_{t} \sim \pi_\theta(\cdot \mid \hat{s}_t, \hat{s}_{t+1})
\end{equation}
The intrinsic reward is then formulated as a sparse binary signal using the indicator function $\mathbb{I}(\cdot)$:
\begin{equation}
    r^i_t = \mathbb{I}(\hat{a}_{t} = a_{t}) = 
    \begin{cases} 
    1, & \text{if } \hat{a}_{t} = a_{t} \\
    0, & \text{otherwise}
    \end{cases}
\end{equation}
where $a_{t}$ denotes the ground-truth action actually executed by the agent in turn $t$. Similiar to Eq.\ref{eq:total_reward}, $r^i_t$ is balanced by weight $\alpha$.

While straightforward, this approach suffers from two critical limitations during policy optimization:
\begin{itemize}
    \item \textbf{Sparsity:} The binary nature of the signal provides zero-gradient information when the prediction is incorrect, which severely hinders efficient gradient estimation for policy optimization.
    \item \textbf{Shortcut Learning:} Since the ground-truth action $a_{t}$ is inherently present in the agent's interaction history, the agent tends to exploit this information leakage by simply retrieving the action from memory (reward hacking), rather than explicitly reasoning about the underlying environmental dynamics.
\end{itemize}

\subsection{Log-Likelihood Scoring}
\label{sec:append_reward_design_log}
This strategy shares the same discriminative scoring mechanism as the Margin-based one, where the policy evaluates candidate actions and outputs a score vector $\mathbf{v} \in \mathbb{R}^{|\mathcal{A}|}$. For consistent implementation across methods, we instruct the model to assign scores within a discrete integer range of $[-5, 5]$. Note that this setting also determines the normalization constants in Eq.~\ref{eq:intrinsic_reward}, where $S_{\text{max}}=5$ and $S_{\text{min}}=-5$.

Unlike the margin-based formulation, this approach treats the scores as unnormalized logits. We first apply a Softmax function to convert $\mathbf{v}$ into a probability distribution $p$:
\begin{equation}
    p(a) = \frac{\exp(\mathbf{v}^{(a)})}{\sum_{a' \in \mathcal{A}} \exp(\mathbf{v}^{(a')})}
\end{equation}
The intrinsic reward is then calculated as the log-probability of the ground-truth action $a_{t}$, centered by the log-probability of a random policy (uniform distribution):
\begin{equation}
    r^i_t = \log p(a_{t}) - \log \left( \frac{1}{|\mathcal{A}|} \right)
\end{equation}
Subtracting $\log(1/|\mathcal{A}|)$ ensures that the reward is positive when the model's confidence exceeds random guessing and negative otherwise. Intuitively, this aligns with the expectation that the signal strength should scale with the extent to which the correct inference deviates from random chance.

While statistically sound, we observe that this formulation leads to training instability due to the \textbf{asymmetric sensitivity} of the logarithmic function. The reward magnitude explodes for low-probability events, overshadowing the signal from correct predictions.
Consider a standard environment with an action space size of $|\mathcal{A}|=4$. The random baseline is $\log(0.25) \approx -1.39$.
\begin{itemize}
    \item If the agent is confident and correct ($p(a_{t}) \approx 0.999$), the reward is bounded: $\log(0.999) - \log(0.25) \approx + 1.38$.
    \item However, if the agent assigns a low probability ($p(a_{t}) \approx 0.001$), the penalty becomes massive: $\log(0.001) - \log(0.25) \approx -5.62$.
\end{itemize}
This property induces large fluctuations with a highly asymmetric range ($r^i_t \in (-\infty, \log |\mathcal{A}|]$). As demonstrated in our ablation study (Fig.~\ref{fig:ablation_reward_weight}), the intrinsic signal functions best as a \textit{curiosity regularizer} rather than a \textit{dominator}. However, the log-likelihood formulation creates a fundamental dilemma for calibrating the weighting coefficient $\alpha$. Consider the previous example where the positive reward is $+1.38$ but the penalty can easily reach $-5.62$ or worse. 
\begin{itemize}
    \item To prevent the massive penalty (dominator) from overriding the extrinsic task reward and causing overly conservative behavior, $\alpha$ must be set to a very small value.
    \item However, heavily scaling down the penalty simultaneously renders the positive signal negligible (e.g., shrinking $+1.38$ to near zero), thereby stripping the signal of its ability to guide the policy.
\end{itemize}
Consequently, it is difficult to find a single $\alpha$ that allows the intrinsic reward to consistently serve as a gentle regularizer across both success and failure cases.

\subsection{Case Study: Comparison of Reward Calculations}
To intuitively illustrate the behavioral differences among the three reward formulations, we provide a concrete case study in Tab.~\ref{tab:reward_calculation_case}. Consider a scenario where the action space is $\mathcal{A} =[\text{Up}, \text{Down}, \text{Left}, \text{Right}]$, and the actual executed action (ground truth) is \texttt{Right}. The agent, however, hallucinates and infers the action was \texttt{Left}.

As shown in the calculations, \textbf{Direct Token Matching} yields a binary zero, providing no gradient information regarding how far off the prediction was. \textbf{Log-Likelihood Scoring} results in a massive penalty ($-8.61$), demonstrating the high-variance and instability issues discussed in Sec.~\ref{sec:append_reward_design_log}. In contrast, our \textbf{Margin-based Scoring} gracefully bounds the penalty to $-1.0$, providing a stable and dense learning signal.

\begin{table}[h]
\centering
\footnotesize
\caption{Case study comparing the three intrinsic reward calculations. The executed ground-truth action is \texttt{Right}, but the agent's retrospection leans towards \texttt{Left}.}
\label{tab:reward_calculation_case}
\resizebox{\textwidth}{!}{
\renewcommand{\arraystretch}{1.5}
\begin{tabular}{p{0.22\textwidth} p{0.45\textwidth} p{0.3\textwidth}}
\toprule
\textbf{Method} & \textbf{Agent Retrospection Output} & \textbf{Calculation Detail \& Final Reward} \\
\midrule
\textbf{Direct Token Matching} & 
\texttt{<retrospection>I think the possible action for causing the environment state transition is: \textbf{Left}. </retrospection>} & 
$r^i = \mathbb{I}(\texttt{Left} == \texttt{Right})$ \newline\newline
\textbf{Result:} $\mathbf{0.00}$ \\
\midrule
\textbf{Log-Likelihood Scoring} \newline ($\mathbf{v}=[-1, -5, 5, -5]$) & 
\texttt{<retrospection>I think the scores of actions in [Up, Down, \textbf{Left}, Right] for causing the environment state transition is: [-1,-5,\textbf{5},-5]. </retrospection>} & 
$p(\texttt{Right}) = \frac{e^{-5}}{e^{-1} + e^{-5} + e^{5} + e^{-5}}$ \newline
$p(\texttt{Right}) \approx 4.53 \times 10^{-5}$ \newline
$r^i = \log(p(\texttt{Right})) - \log(0.25)$ \newline
$r^i = -10.00 - (-1.39)$ \newline
\textbf{Result:} $\mathbf{-8.61}$ (Massive Penalty) \\
\midrule
\textbf{Margin-based Scoring} \newline (\textbf{Ours}) \newline ($\mathbf{v}=[-1, -5, 5, -5]$) & 
\texttt{<retrospection>I think the scores of actions in [Up, Down, \textbf{Left}, Right] for causing the environment state transition is:[-1,-5,\textbf{5},-5]. </retrospection>} & 
$v(\text{Right}) = -5$ \newline
$\max_{a \neq \texttt{Right}} v(a) = v(\texttt{Left}) = 5$ \newline
$S_{\max} = 5, S_{\min} = -5$ \newline
$r^i = \frac{-5 - 5}{5 - (-5)} = \frac{-10}{10}$ \newline
\textbf{Result:} $\mathbf{-1.00}$ (Bounded Penalty) \\
\bottomrule
\end{tabular}
}
\end{table}

\section{Details of Advantage Estimation}
\label{sec:append_advantage_estimation}
As discussed in Sec.~\ref{sec:policy_optimization}, our composite reward $r_t$ (which importantly encompasses the intrinsic Self-Consistency Reward $r_t^i$) is evaluated at the macro-level of environmental turns. However, optimizing the VLM policy via PPO requires fine-grained advantage estimates at the micro-level of individual tokens. To bridge this granularity gap, we adopt the Bi-Level General Advantage Estimation (GAE) protocol introduced by VAGEN~\cite{wang2025vagen}. 

For completeness and to clarify our specific implementation within the Retrospective World Modeling (RWM) framework, we formally describe this two-stage credit assignment process below:

\textbf{Turn-Level Advantage Estimation.}
We first calculate the advantage for each environmental turn based on the macro-level reward $r_t$. Let $\bar{\tau}_{\le a_{t}}$ denote the full token prefix of the trajectory up to and including the generated action at turn $t$. Using the critic network $V_\phi$, we compute the turn-level TD-error $\delta^{\mathrm{turn}}_t$ and subsequently the turn-level advantage $A^{\mathrm{turn}}_t$ via standard backward propagation across turns:
    \begin{equation}
        \delta^{\mathrm{turn}}_t = r_t + \gamma_{\mathrm{turn}} V_\phi(\bar{\tau}_ {\le a_{t+1}}) - V_\phi(\bar{\tau}_{\le a_t})
    \end{equation}
    \begin{equation}
        A^{\mathrm{turn}}_t = \delta^{\mathrm{turn}}_t + \gamma_{\mathrm{turn}} \lambda_{\mathrm{turn}} A^{\mathrm{turn}}_{t+1}
    \end{equation}
where $\gamma_{\mathrm{turn}}$ and $\lambda_{\mathrm{turn}}$ are the discount factor and the GAE decay parameter at the turn level, respectively.

\textbf{Token-Level Advantage Estimation.}
After deriving the turn-level advantages, we perform an inner GAE loop for the generated tokens within each specific action $a_t$. Let $\delta^{\mathrm{token}}_{t,i}$ represent the intrinsic token-level TD-error for the $i$-th token, which is typically derived from the KL divergence penalty against the reference model to prevent policy degradation. 
    
Crucially, to inject the episodic semantic feedback into token generation, the backward pass for the token-level advantages is anchored by the turn-level advantage. Specifically, the aggregated signal $A^{\mathrm{turn}}_t$ is incorporated into the advantage of the final generated token of action $a_t$. Standard GAE then propagates this fused signal backward through all preceding tokens within that specific turn:
    \begin{equation}
        A^{\mathrm{token}}_{t,i} = \delta^{\mathrm{token}}_{t,i} + \gamma_{\mathrm{token}}\lambda_{\mathrm{token}} A^{\mathrm{token}}_{t,i+1}
    \label{eq:token_advantage}
    \end{equation}
Token advantage $A^{\mathrm{token}}_{t,i+1}$ is used in the policy optimization, as illustrated in Eq.~\ref{eq:PPO}. By doing so, the dense self-consistency verification signal formulated at the turn level is successfully distributed to optimize the fine-grained token generation process.

\section{Experimental Details}
\label{sec:append_experiment_details}
We use \texttt{Qwen2.5-VL-3B}~\cite{bai2023qwenvlversatilevisionlanguagemodel} as the backbone. During training, the global batch size is set to 128, with a learning rate of $1 \times 10^{-6}$ for the actor and $1 \times 10^{-5}$ for the critic.
For evaluation, we set the generation temperature to 1.0.
Results are averaged over 3 runs on 128 diverse test cases to ensure statistical significance.
top-p 1.0, and max new tokens 1024. 
We repeat 3 evaluations on 128 test cases generated from the environment and calculate the average to mitigate randomness.
For agent's interaction with environment, the maximum turn number is set to 4. 
To support retrospective reasoning, the interaction trajectory $\tau_t$ is structured as: 
{\small\ttfamily
<think><Obs>$\hat{s}_t$</Obs><Retro>$\hat{a}_{t-1}$</Retro><Rea>
$a^{\prime}_t$</Rea><Pred>$s^{\prime}_{t+1}$</Pred></think><Ans>$a_t$</Ans>}, 
where \texttt{<Obs>}, \texttt{<Retro>}, \texttt{<Rea>}, \texttt{<Pred>} and \texttt{<Ans>} are abbreviations for observation, retrospection, reasoning, prediction and final answer. 
All experiments were conducted on 4$\times$ NVIDIA H800 (80GB) GPUs.
\subsection{Baselines}
\label{sec:append_setting_baselines}
To comprehensively evaluate the effectiveness of our Retrospective World Modeling (RWM) paradigm, we compare it against a spectrum of reinforcement learning baselines. These baselines are carefully selected to isolate the contributions of intermediate reasoning, structured world modeling, and advantage estimation techniques.

\textbf{Vanilla-PPO}~\cite{ragen}. This is the most fundamental RL baseline, which directly maps observations to executable actions. The agent is trained purely on sparse environmental rewards at the end of the trajectory using standard Token-Level GAE. It serves to demonstrate the baseline performance of the VLM when it acts purely reactively.

\textbf{ReAct-RL}~\cite{yao2023react}. Building upon the direct-action approach, this baseline incorporates the ReAct paradigm by prompting the agent to generate free-form, natural language reasoning traces (``observe-reason-act'') before committing to an action. Like Vanilla-PPO, it is optimized using Token-Level GAE with sparse task rewards. This baseline tests whether unstructured chain-of-thought generation is sufficient for complex agentic tasks without explicit world modeling constraints.

\textbf{VAGEN-Base}~\cite{wang2025vagen}. As a more challenging exploration baseline, VAGEN-Base introduces structured \textit{prospective} world modeling, prompting the agent to explicitly generate state estimations and predict future transitions. However, it strips away any dense semantic feedback, relying entirely on sparse task success and basic formatting rewards. For credit assignment, it falls back to standard Token-Level GAE, where the sparse reward is anchored to the final token of the trajectory and propagated backward across all action tokens without explicit turn boundaries. This setup rigorously tests whether prospective world modeling capabilities can naturally emerge from pure trial-and-error interactions without explicit guidance.

\textbf{VAGEN-Full}~\cite{wang2025vagen}. Representing the upper bound of the prospective reasoning paradigm, VAGEN-Full provides dense supervision to the agent's internal world model. It calculates a comprehensive reasoning reward using an external LLM-as-a-Judge to directly evaluate the factual correctness of the agent's prospective state estimations and predictions. Crucially, to resolve the temporal credit assignment bottleneck inherent in multi-turn RL, it implements the Bi-Level GAE mechanism. By computing a macro-scale turn-level advantage ($A_t^{turn}$) first and injecting it as a terminal target into the micro-scale token-level inner-GAE, this setup ensures that episodic success is explicitly credited to the specific reasoning tokens that justified the action.

\vspace{1em}
\noindent\textbf{Summary of Method Configurations.} 
To provide a clear conceptual boundary between the prospective VAGEN framework and our proposed retrospective RWM framework, we summarize their structural and algorithmic differences. While both frameworks share the concept of explicitly structuring the VLM's internal thoughts, they fundamentally diverge in their reasoning direction (Prospective vs. Retrospective). These distinctions are detailed in Table~\ref{tab:setting_comparison}.

\begin{table}[h]
\centering
\caption{Comparison of training configurations across VAGEN and RWM variants.}
\label{tab:setting_comparison}
\resizebox{\textwidth}{!}{
\begin{tabular}{l l l l}
\toprule
\textbf{Setting} & \textbf{Reasoning Direction} & \textbf{Credit Assignment} & \textbf{Dense Reward Components} \\
\midrule
VAGEN-Base & Prospective Only & Token-Level GAE & None (Sparse Task and Format Reward Only) \\
VAGEN-Full & Prospective Only & Bi-Level GAE    & LLM-as-a-Judge Reward \\
\midrule
RWM-Base   & Prospective + Retrospective & Token-Level GAE & None (Sparse Task and Format Reward Only) \\
RWM-Full   & Prospective + Retrospective & Bi-Level GAE    & LLM-as-a-Judge + \textbf{Self-Consistency Reward} \\
\bottomrule
\end{tabular}
}
\end{table}
\subsection{Environments}
\label{sec:append_setting_env}
In this section, we introduce the information and the parameters of the environments that we use in the experiments. The selected tasks span a wide range of challenges, encompassing reasoning over both 2D symbolic layouts and 3D photorealistic scenes, and demonstrate adaptability to diverse embodiment settings, including household agents and game environments.

\textbf{Sokoban} is a classic grid-based puzzle in which the agent is required to push all boxes to target locations. The environment is a 2D grid with a discrete action space consisting of four movements (up, down, left, right). For Sokoban, we use a 6 × 6 grid as the standard setting for all models. Meanwhile, we construct a Hard setting of Sokoban to probe generalization on more demanding tasks, by increasing the map dimensions by 2. Detailed information is listed in Tab.\ref{tab:append_info_sokoban_env}.
\begin{table}[h]
    \centering
    \caption{Information about Sokoban Environment.}
    \label{tab:append_info_sokoban_env}
    \renewcommand{\arraystretch}{1.2} 
    \begin{tabular}{l|l}
        \toprule
        \textbf{Name} & \textbf{Value} \\
        \hline
        room size & $6\times6$, $8\times8$ \\
        \hline
        num of boxes & 1 \\
        \hline
        maximum turn number & 4 \\
        \hline
        action space & Up, Down, Left, Right \\
        \hline
        format reward & 0.5 \\
        \hline
        step penalty & $-0.1$ \\
        \hline
        success reward & 10  \\
        \hline
        intrinsic reward weight $\alpha$ & 0.1 \\
        \hline
        example & \makecell[l]{\includegraphics[width=0.1\textwidth]{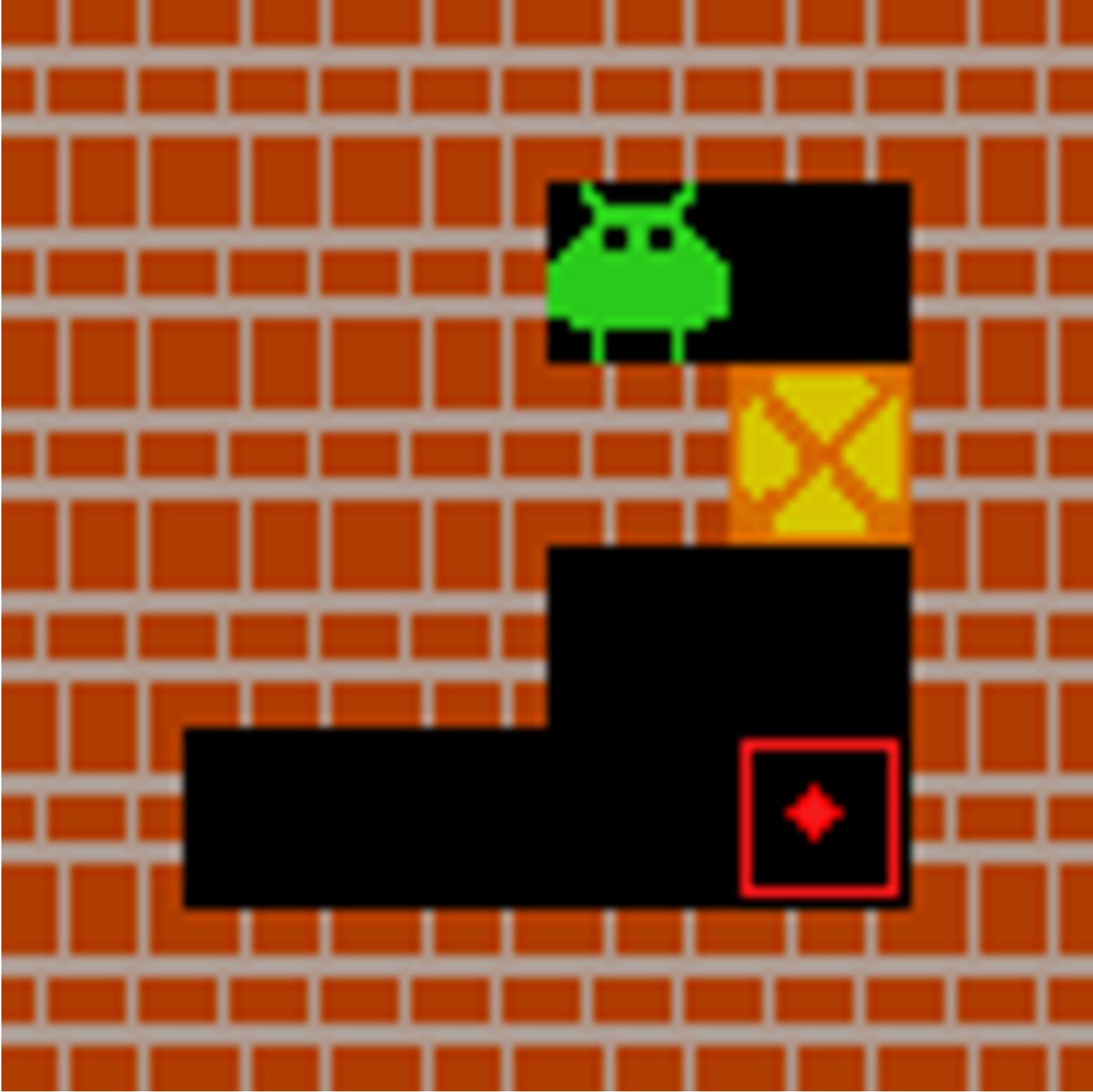}} \\
        \bottomrule
    \end{tabular}
\end{table}


\textbf{FrozenLake} is a 2D grid-based map in which the agent need to reach a goal while avoiding holes. The visual state and discrete action space are similar to Sokoban. The default slippery dynamics are disabled to ensure deterministic transitions. For FrozenLake, we use 4 × 4 grid for all models. Detailed information is listed in Tab.\ref{tab:append_info_frozenlake_env}.

\begin{table}[!h]
    \centering
    \caption{Information about FrozenLake Environment.}
    \label{tab:append_info_frozenlake_env}
    \renewcommand{\arraystretch}{1.2} 
    \begin{tabular}{l|l}
        \toprule
        \textbf{Name} & \textbf{Value} \\
        \hline
        room size & $4\times4$ \\
        \hline
        is slippery & False \\
        \hline
        maximum turn number & 4 \\
        \hline
        action space & Up, Down, Left, Right \\
        \hline
        format reward & 0.5 \\
        \hline
        step penalty & $-0.1$ \\
        \hline
        success reward & 10  \\
        \hline
        intrinsic reward weight $\alpha$ & 0.05 \\
        \hline
        example & \makecell[l]{\includegraphics[width=0.1\textwidth]{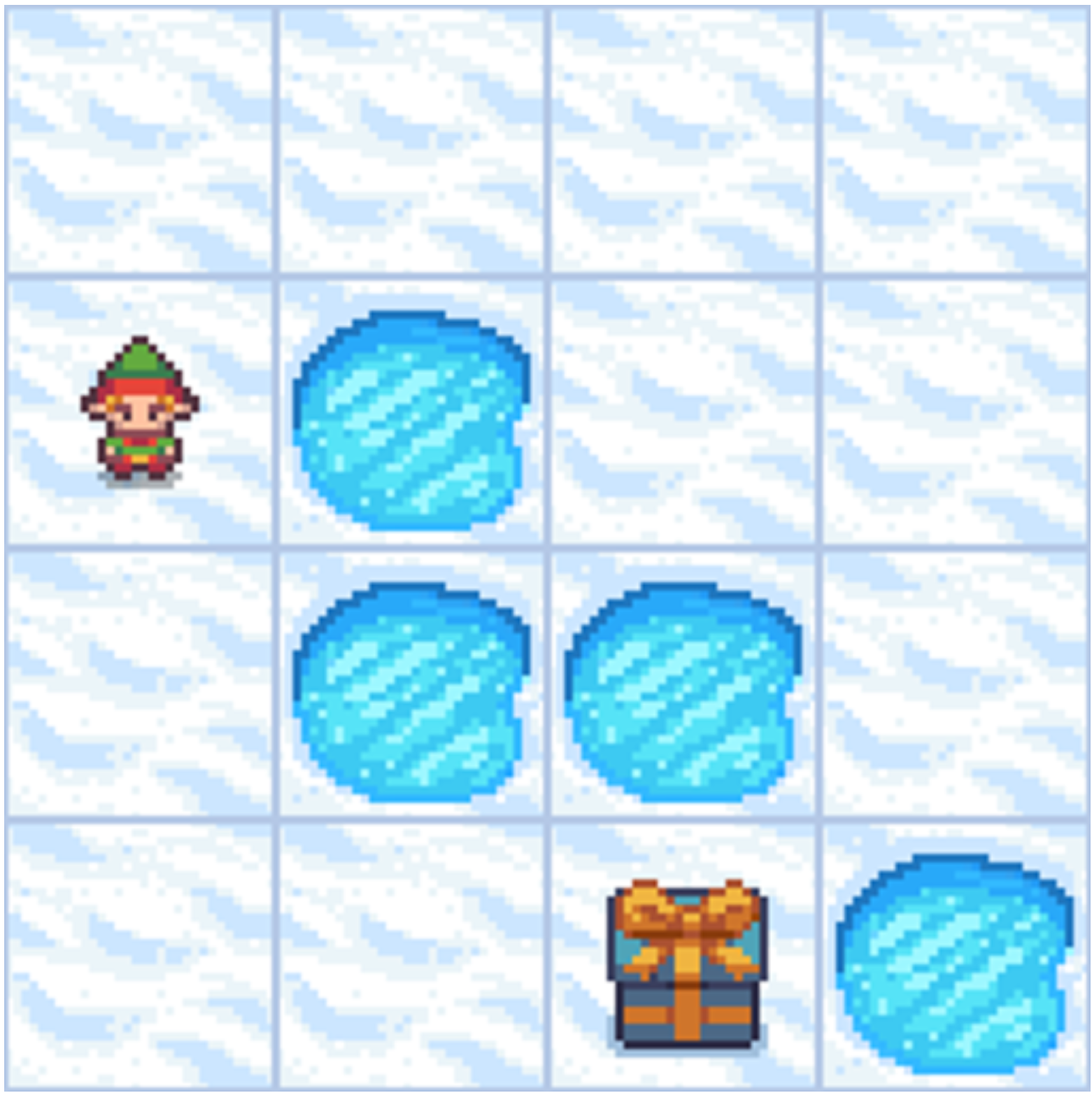}} \\
        \bottomrule
    \end{tabular}
\end{table}

\textbf{Navigation} is a 3D embodied navigation task in which the agent follows instructions to locate a target object. The agent perceives the environment from a first-person perspective and interacts through a discrete action space (e.g., \texttt{MoveAhead}). For Navigation, we evaluate on two eval sets: base and common sense. Detailed information is listed in Tab.\ref{tab:append_info_navigation_env}.

\begin{table}[!h]
    \centering
    \caption{Information about Navigation Environment.}
    \label{tab:append_info_navigation_env}
    \renewcommand{\arraystretch}{1.2} 
    \begin{tabular}{l|l}
        \toprule
        \textbf{Name} & \textbf{Value} \\
        \hline
        resolution & 255 \\
        \hline
        success threshold & 1.5 \\
        \hline
        step distance & 0.5 \\
        \hline
        action space & \makecell[l]{MoveAhead, MoveBack,\\ MoveRight, MoveLeft,\\ RotateRight, RotateLeft,\\ LookUp, LookDown}\\
        \hline
        format reward & 0.5 \\
        \hline
        step penalty & $-0.1$ \\
        \hline
        success reward & 10  \\
        \hline
        intrinsic reward weight $\alpha$ & 0.05 \\
        \hline
        example & \makecell[l]{\includegraphics[width=0.1\textwidth]{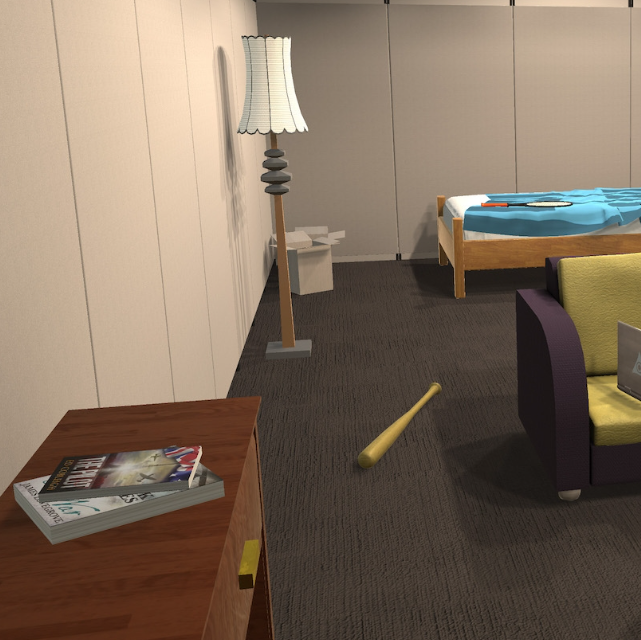}} \\
        \bottomrule
    \end{tabular}
\end{table}

\textbf{ManiSkill} is also a 3D embodied manipulation benchmark where an agent controls a Panda robotic arm under stochastic dynamics. It features a hybrid action space that combines discrete action types with continuous parameters (e.g., \texttt{pick(x, y, z)}). Given third-person 3D observations, the agent must perform fine-grained visual grounding to map objects to precise coordinates, enabling accurate and physically executable interactions. Detailed information is listed in Tab.\ref{tab:append_info_maniskill_env}. Since actions in this environment consist of two components, including a discrete action type and continuous spatial parameters, we adapt SCR to jointly infer both the action type $\hat{u}_{t}$ (i.e., pick, place, or push) and the continuous parameters $\hat{p}_{t}$ (i.e., spatial coordinates). The modified SCR enforces a strict type match via an indicator function $\mathbb{I}$, followed by a normalized spatial penalty (with $D_{\max}$ denoting the maximum distance):
\begin{equation}
\begin{aligned}
r^i_t &= \mathbb{I}(\hat{u}_{t} = u_{t}) \cdot \left( 1 - \frac{\|\hat{p}_{t} - p_{t}\|_2}{D_{\max}} \right)
\end{aligned}
\end{equation}

\begin{table}[!h]
    \centering
    \caption{Information about ManiSkill Environment.}
    \label{tab:append_info_maniskill_env}
    \renewcommand{\arraystretch}{1.2} 
    \begin{tabular}{l|l}
        \toprule
        \textbf{Name} & \textbf{Value} \\
        \hline
        action space & \makecell[l]{pick(x, y, z),\\ place(x, y, z),\\ push(x1, y1, z1, x2, y2, z2)}\\
        \hline
        format reward & 0.5 \\
        \hline
        step penalty & $-0.1$ \\
        \hline
        success reward & 10  \\
        \hline
        intrinsic reward weight $\alpha$ & 0.05 \\
        \hline
        example & \makecell[l]{\includegraphics[width=0.1\textwidth]{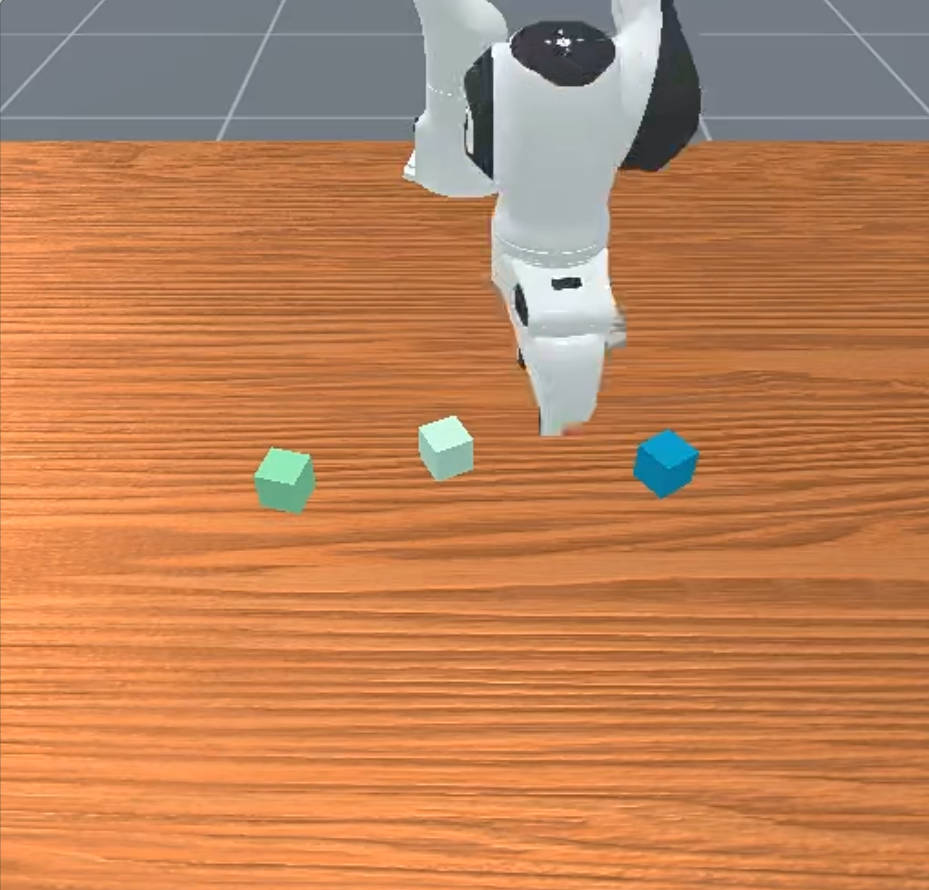}} \\
        \bottomrule
    \end{tabular}
\end{table}

\section{Discussion and Future Direction}
\label{sec:discussion}
\begin{figure}[h]
\centering
\includegraphics[width = 0.49\textwidth]{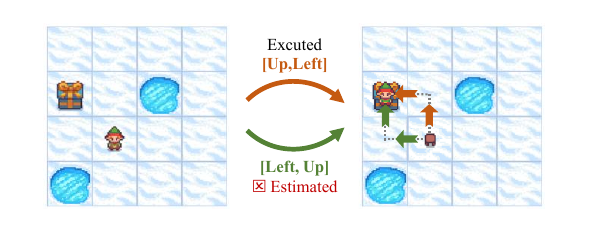}
\caption{Case Study: Causal Ambiguity in Multi-Action Turns. Distinct action sequences can induce identical state transitions, posing challenges to the horizon of causality.}
\label{fig:horizon}
\end{figure}

\paragraph{The Horizon of Causality.}
While RWM demonstrates strong efficacy in standard interactive settings, our analysis reveals a fundamental challenge of \textit{causal ambiguity}.
Our framework assumes that the transition $(s_t, s_{t+1})$ contains sufficient information to infer the underlying action.
Importantly, we also account for more general settings where the temporal horizon between observations expands, such as multi-action turns in which an agent executes a sequence $\{a_1, \ldots, a_k\}$ before receiving $o_{t+1}$. In these cases, the correspondence between state changes and actions naturally becomes many-to-one.
Consider a simple FrozenLake scenario in Fig.~\ref{fig:horizon}: the sequences \texttt{<Up, Left>} and \texttt{<Left, Up>} may lead to an identical final state configuration. If the agent executes the former while the retrospective world modeling infers the latter, the resulting intrinsic reward would incorrectly penalize a valid action, yielding a false negative signal, since the executed action is already fixed and unambiguous.
This underscores a critical insight for the field of world modeling: \textit{dense observation is a prerequisite for accurate causal grounding}.
When observations are sparse, the ground truth becomes ambiguous.
To extend the horizon of causality, future work in world modeling must address this by moving beyond single-trajectory lookback to set-based retrospection, where the reward acknowledges the set of all physically valid causal paths rather than overfitting to a single deterministic history.

\paragraph{Limitation.}
While RWM significantly improves agent reasoning and decision-making, we acknowledge two primary limitations. First, in environments with \textit{extreme stochastic dynamics}, the inverse transition mapping can become ambiguous (i.e., multiple distinct actions or random environmental perturbations may yield the exact same next observation). This causal ambiguity could introduce noise into the Self-Consistency Reward, which may require more advanced probabilistic modeling in future work. Second, scaling RWM to \textit{extremely long-horizon tasks} currently poses practical hardware challenges. Training multi-turn VLM agents via RL is highly memory-intensive; extending the interaction to very long turns naturally encounters Out-Of-Memory (OOM) bottlenecks. Future engineering optimizations in memory-efficient RL (e.g., context compression or KV-cache offloading) should be explored to deploy this paradigm in longer, more complex scenarios.
\newpage
\section{Case Study}

We report some cases of the agent trained with our RWM, as shown in Tab.\ref{tab:case_study_sokoban} and Tab.\ref{tab:case_study_frozonlake}.

\begin{table}[h]
\centering
\caption{Case study for our agent in a Sokoban environment.}
\label{tab:case_study_sokoban}
\renewcommand{\arraystretch}{1.2} 
\begin{tabular}{l|l}
\toprule
\textbf{Name} & \textbf{Value} \\
\hline
state\_0 & \makecell[l]{\includegraphics[width=0.1\textwidth]{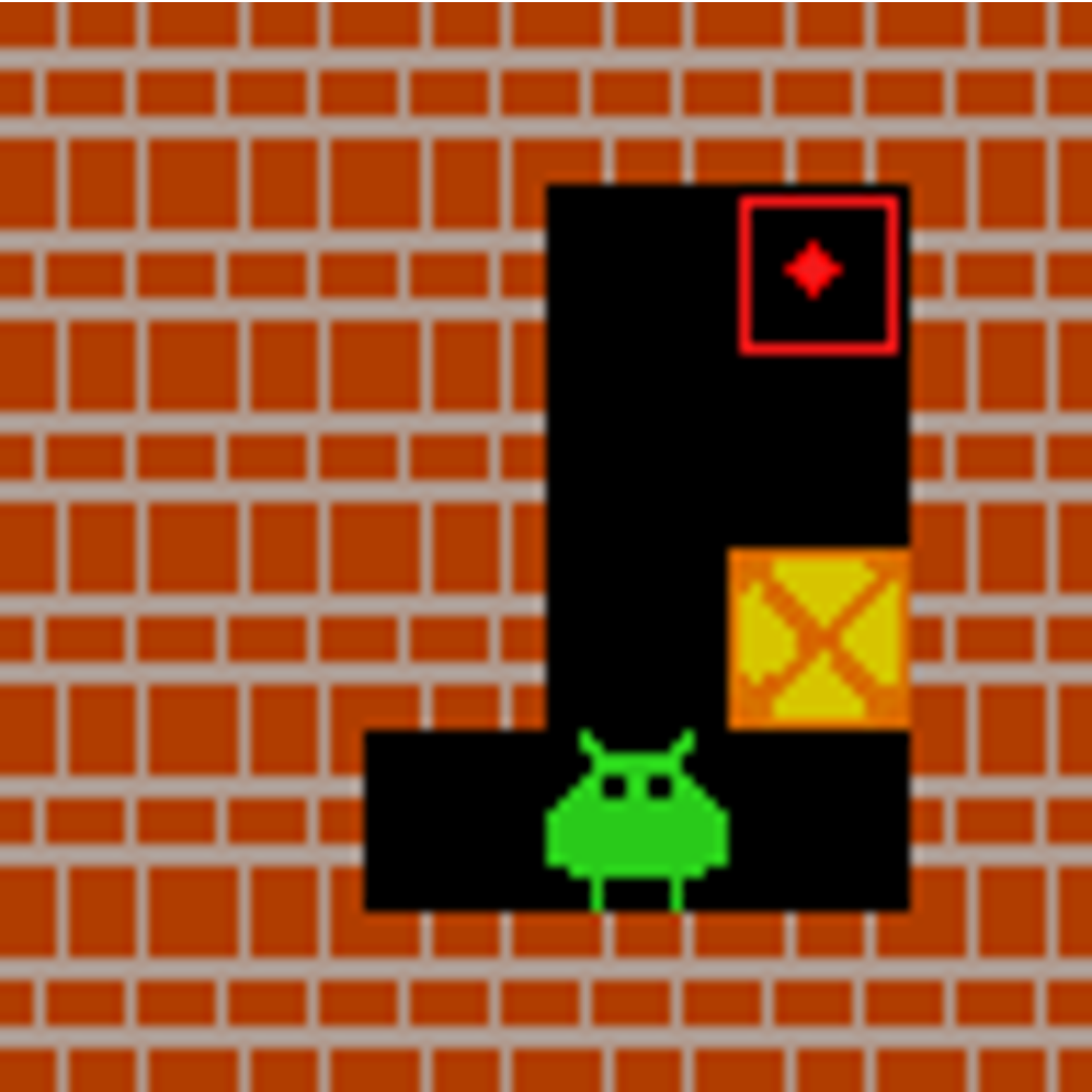} (initial state)} \\
\hline
agent & \makecell[l]{
\textless think\textgreater \\
\textcolor{darkgreen}{\textless observation\textgreater The box is above and to the right of the player. The target is above the box. \textless/observation\textgreater }\\
\textcolor{blue}{\textless reasoning\textgreater The player needs to go to the right first. \textless/reasoning\textgreater }\\
\textcolor{darkorange}{\textless prediction\textgreater The player will be below the box. \textless/prediction\textgreater }\\
\textless/think\textgreater \\
\textless answer\textgreater Right \textless/answer\textgreater\\} \\
\hline
state\_1 & \makecell[l]{\includegraphics[width=0.1\textwidth]{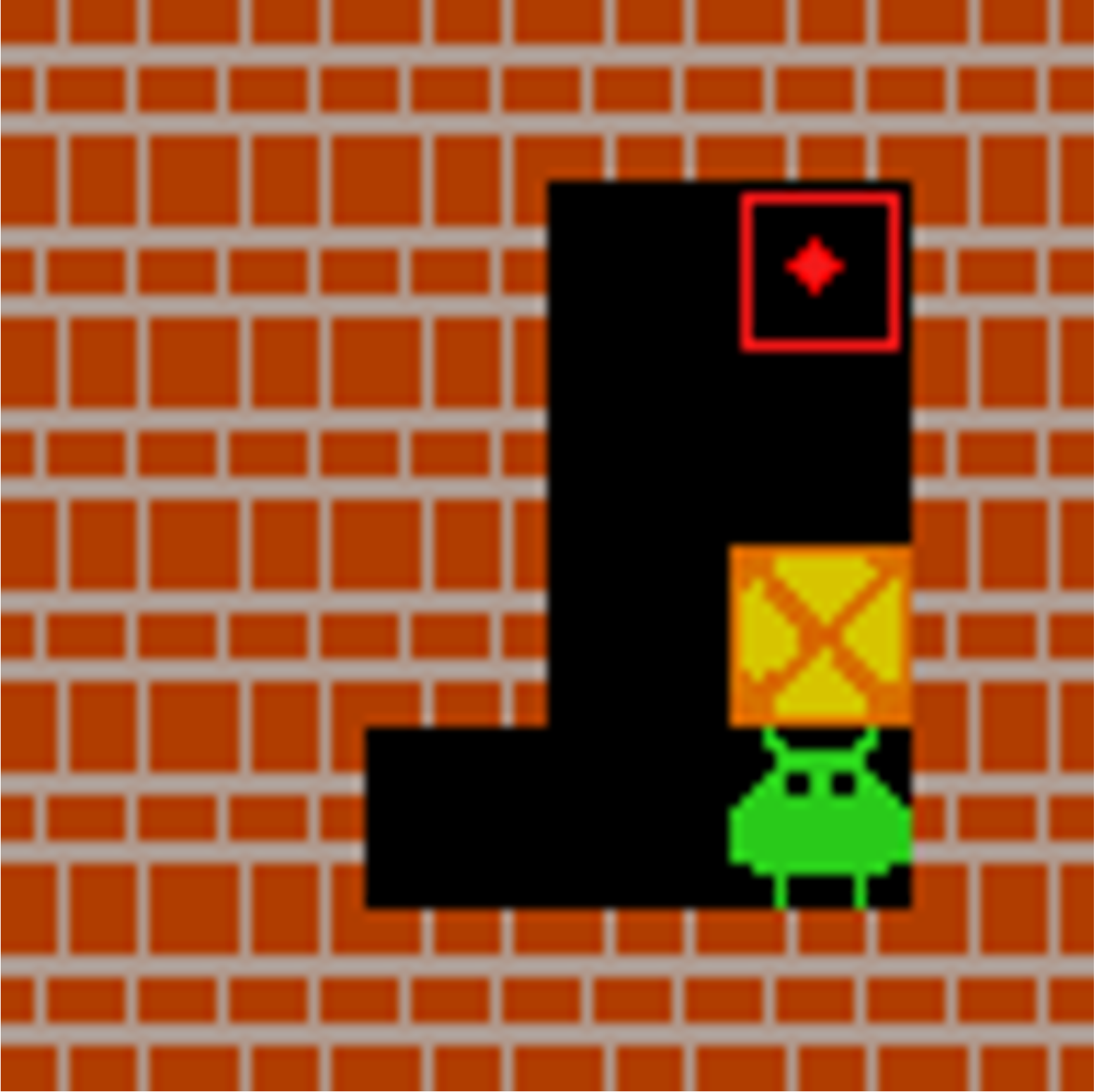}} \\
\hline
agent & \makecell[l]{
\textless think\textgreater \\
\textcolor{darkgreen}{\textless observation\textgreater The box is directly above the player and the target is above the box. \textless/observation\textgreater }\\
\textcolor{red}{\textless retrospection\textgreater I think the scores of actions in [Up, Down, Left, Right] for causing the environment} \\\textcolor{red}{state transition is: [-1,0,0,5]. \textless/retrospection\textgreater }\\
\textcolor{blue}{\textless reasoning\textgreater The player needs to go up to push the box. \textless/reasoning\textgreater }\\
\textcolor{darkorange}{\textless prediction\textgreater The player will be below the box, the target will be above the box. \textless/prediction\textgreater }\\
\textless/think\textgreater \\
\textless answer\textgreater Up \textless/answer\textgreater\\} \\
\hline
state\_2 & \makecell[l]{\includegraphics[width=0.1\textwidth]{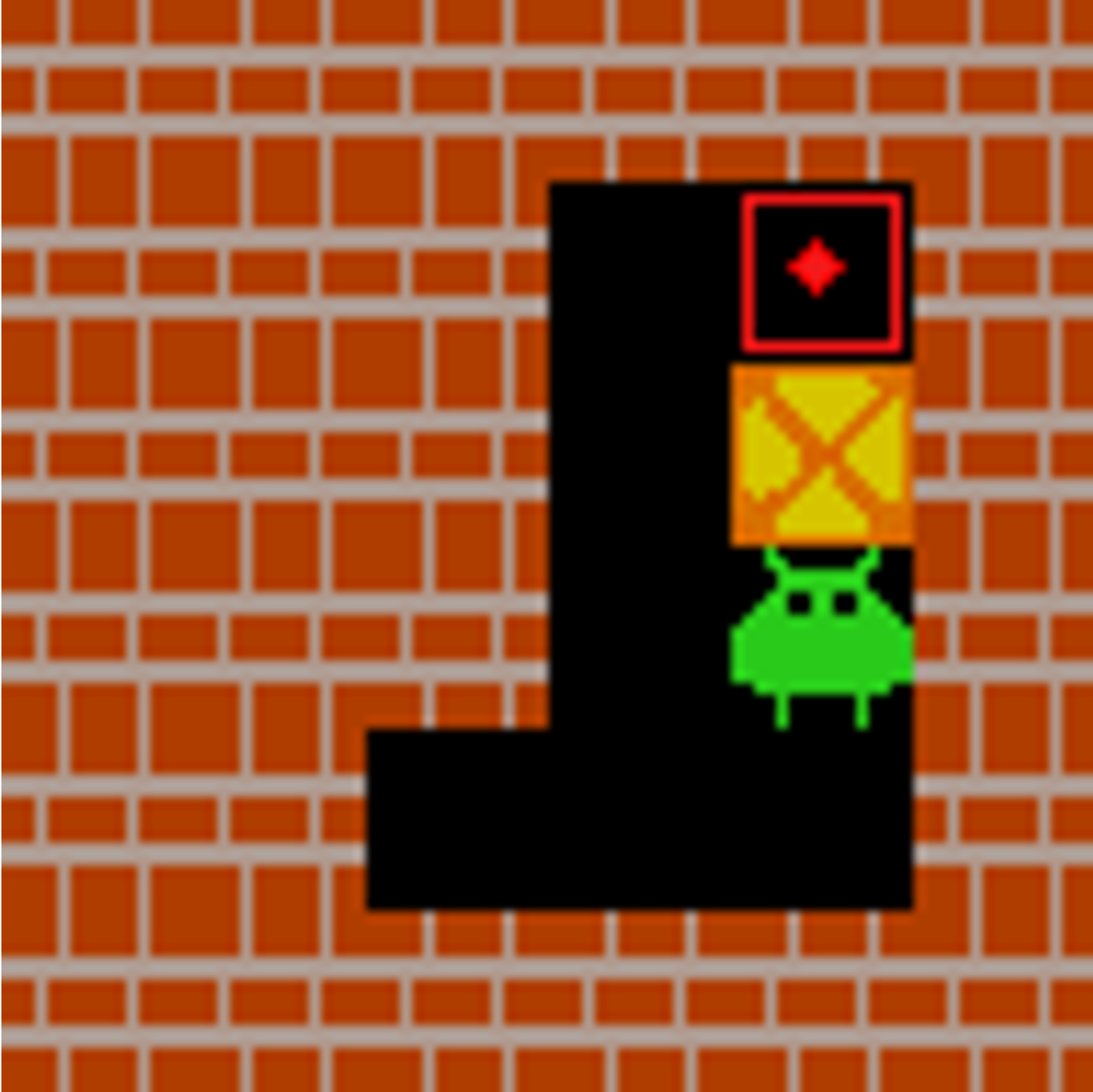}} \\
\hline
agent & \makecell[l]{
\textless think\textgreater \\
\textcolor{darkgreen}{\textless observation\textgreater The box is above the player and the target is directly above the box. \textless/observation\textgreater }\\
\textcolor{red}{\textless retrospection\textgreater I think the scores of actions in [Up, Down, Left, Right] for causing the environment }\\ \textcolor{red}{state transition is: [5,0,-5,-5]. \textless/retrospection\textgreater }\\
\textcolor{blue}{\textless reasoning\textgreater The player needs to go up to push the box up to the target. \textless/reasoning\textgreater }\\
\textcolor{darkorange}{\textless prediction\textgreater The player will be below the box, the box will be at the same position of the target. \textless/prediction\textgreater }\\
\textless/think\textgreater \\
\textless answer\textgreater Up \textless/answer\textgreater\\} \\
\hline
state\_3 & \makecell[l]{\includegraphics[width=0.1\textwidth]{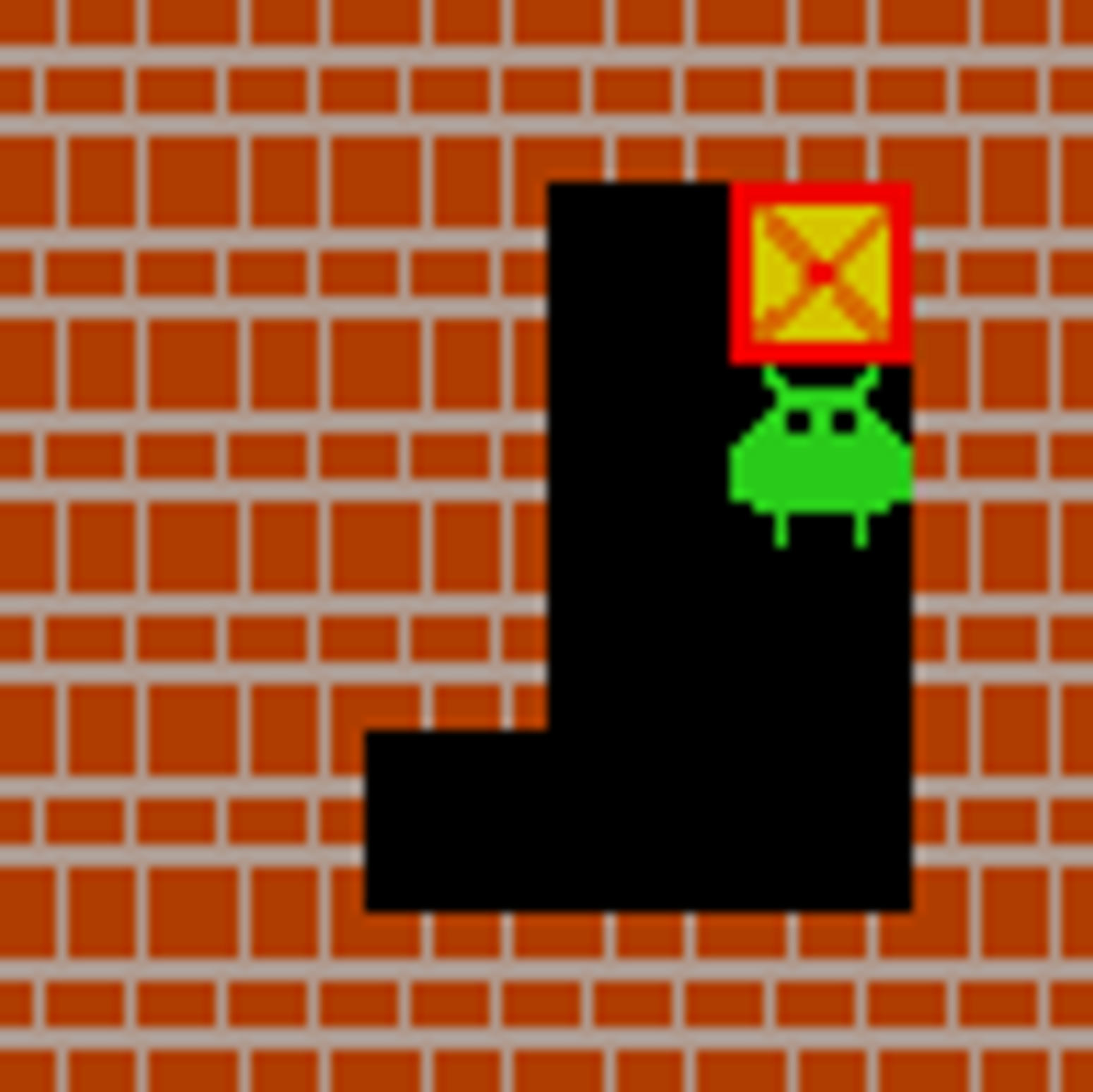} (success)} \\
\bottomrule
\end{tabular}
\end{table}
\clearpage
\newpage

\begin{table}[h]
\centering
\caption{Case study for our agent in a FrozenLake environment.}
\label{tab:case_study_frozonlake}
\renewcommand{\arraystretch}{1.2} 
\begin{tabular}{l|l}
\toprule
\textbf{Name} & \textbf{Value} \\
\hline
state\_0 & \makecell[l]{\includegraphics[width=0.1\textwidth]{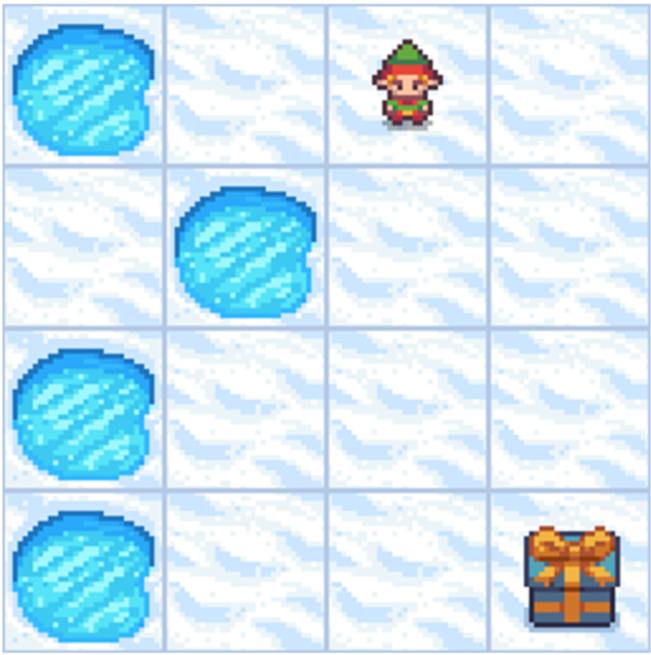} (initial state)} \\
\hline
agent & \makecell[l]{
\textless think\textgreater \\
\textcolor{darkgreen}{\textless observation\textgreater The player is to the left of and above the goal. There is a hole below and to the left of } \\\textcolor{darkgreen}{the player. \textless/observation\textgreater }\\
\textcolor{blue}{\textless reasoning\textgreater I should go right first.  \textless/reasoning\textgreater }\\
\textcolor{darkorange}{\textless prediction\textgreater The player will be above of the goal. \textless/prediction\textgreater }\\
\textless/think\textgreater \\
\textless answer\textgreater Right \textless/answer\textgreater\\} \\
\hline
state\_1 & \makecell[l]{\includegraphics[width=0.1\textwidth]{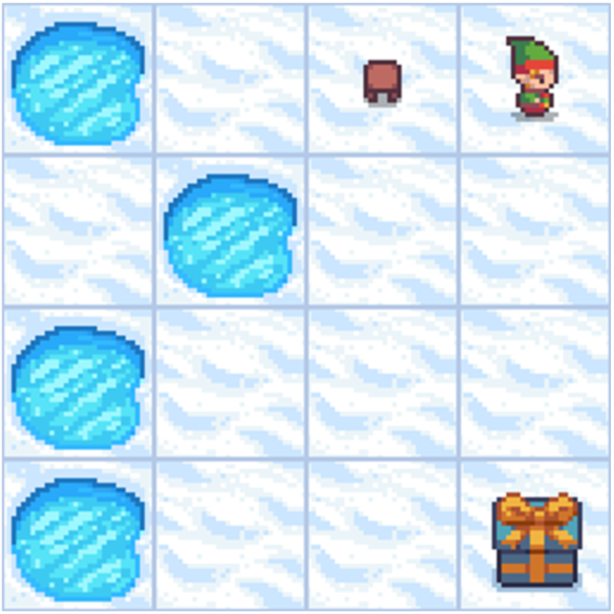}} \\
\hline
agent & \makecell[l]{
\textless think\textgreater \\
\textcolor{darkgreen}{\textless observation\textgreater The player is above the goal. \textless/observation\textgreater }\\
\textcolor{red}{\textless retrospection\textgreater I think the scores for actions [Up, Down, Left, Right] are: [0,-5,-5, 5]. \textless/retrospection\textgreater }\\
\textcolor{blue}{\textless reasoning\textgreater The goal is below me, so I should go down first. \textless/reasoning\textgreater }\\
\textcolor{darkorange}{\textless prediction\textgreater The player will be above of the goal. \textless/prediction\textgreater }\\
\textless/think\textgreater \\
\textless answer\textgreater Down \textless/answer\textgreater\\} \\
\hline
state\_2 & \makecell[l]{\includegraphics[width=0.1\textwidth]{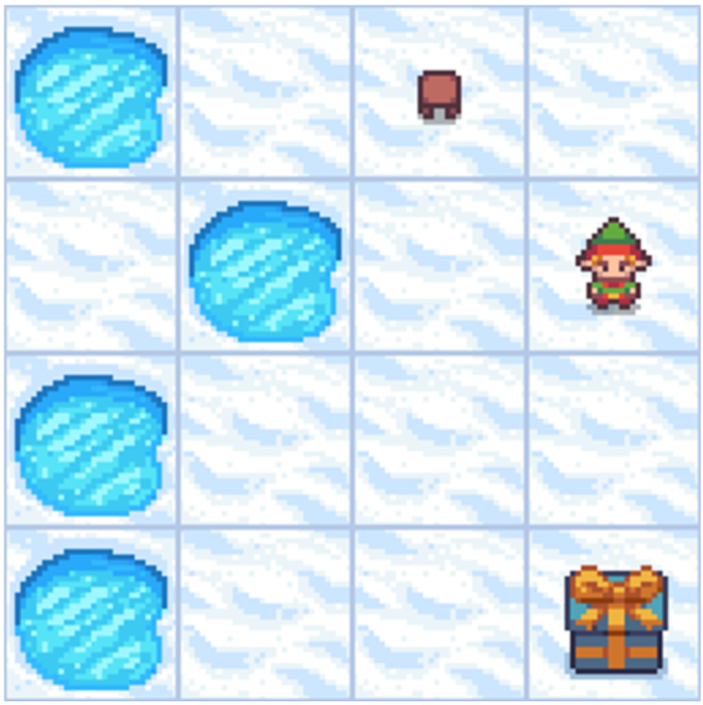}} \\
\hline
agent & \makecell[l]{
\textless think\textgreater \\
\textcolor{darkgreen}{\textless observation\textgreater The player is above the goal. \textless/observation\textgreater }\\
\textcolor{red}{\textless retrospection\textgreater I think the scores for actions [Up, Down, Left, Right] are: [-5 ,5,-1, -5]. \textless/retrospection\textgreater }\\
\textcolor{blue}{\textless reasoning\textgreater The goal is below me, so I should go down. \textless/reasoning\textgreater }\\
\textcolor{darkorange}{\textless prediction\textgreater The player will be above of the goal. \textless/prediction\textgreater }\\
\textless/think\textgreater \\
\textless answer\textgreater Down \textless/answer\textgreater\\} \\
\hline
state\_3 & \makecell[l]{\includegraphics[width=0.1\textwidth]{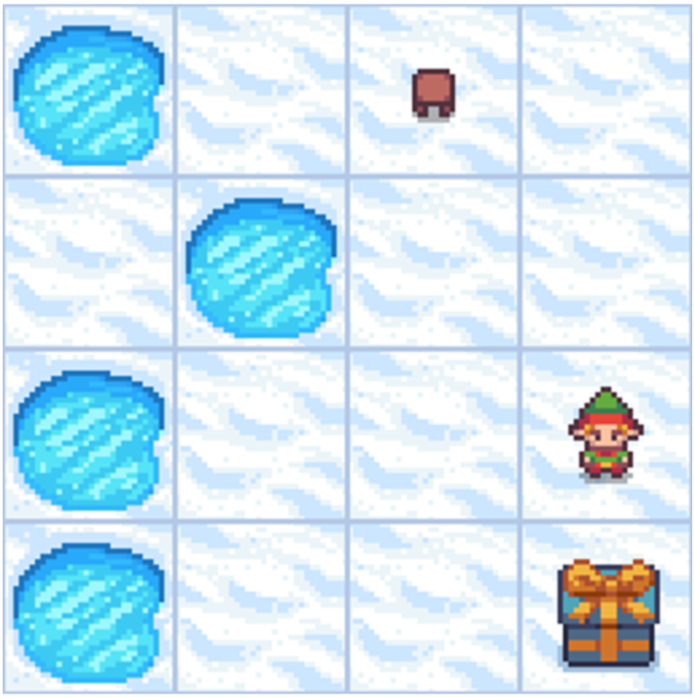}} \\
\hline
agent & \makecell[l]{
\textless think\textgreater \\
\textcolor{darkgreen}{\textless observation\textgreater The goal is below the player. \textless/observation\textgreater }\\
\textcolor{red}{\textless retrospection\textgreater I think the scores for actions [Up, Down, Left, Right] are: [-5,5,-5,-5]. \textless/retrospection\textgreater }\\
\textcolor{blue}{\textless reasoning\textgreater The goal is below me, so I should go down. \textless/reasoning\textgreater }\\
\textcolor{darkorange}{\textless prediction\textgreater The player will be at the same position of the goal. \textless/prediction\textgreater }\\
\textless/think\textgreater \\
\textless answer\textgreater Down \textless/answer\textgreater\\} \\
\hline
state\_4 & \makecell[l]{\includegraphics[width=0.1\textwidth]{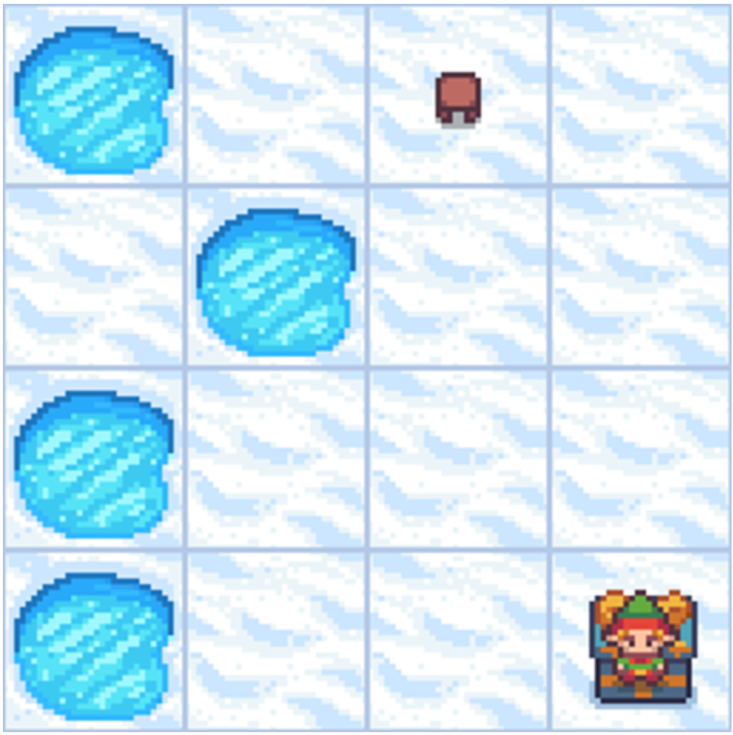} (success)} \\
\bottomrule
\end{tabular}
\end{table}

\clearpage
\newpage
\section{Prompts}
We report the prompts used in the training process, as shown in Tab.~\ref{prompt:prompt_sokoban}, Tab.~\ref{prompt:prompt_frozenlake}, and Tab.~\ref{prompt:prompt_navigation}.\\
\\
\captionof{table}{Prompt for Sokoban.}
\label{prompt:prompt_sokoban}
\begin{tcolorbox}[
  enhanced,                 
  breakable,                
  width=\linewidth,
  colback=gray!10,
  colframe=gray!50,
  boxrule=0.4pt,
  arc=2pt,
  left=6pt,
  right=6pt,
  top=6pt,
  bottom=6pt,
  fonttitle=\itshape\small\color{black},
  fontupper=\small\ttfamily
]
\ttfamily
\textbf{Prompt for Sokoban}\\
-------------------\\
(Initial Turn)\\
-------------------\\
You are a Sokoban solver.\\
Your goal is to push all boxes onto targets.\\

Rules:\\
1. Push boxes (can't pull).\\
2. Avoid walls.\\

Actions you can take: Up, Down, Left, Right.\\
You should take 1 action at a time.\\

Initial Observation:\\
<image>\\

Decide your next action.\\
You should take 1 action at a time.\\

You should first give the description of your observation, then your reasoning, then predict the next state, and finally your answer.\\

For the content you provide within the `<observation>` and `<prediction>` tags, you must strictly describe the relative position of the `target` and any visible `box` objects **relative to the player**. Your description/prediction must include **both** a vertical and a horizontal directional relationship for each object. Use ONLY the terms `above`, `below`, `left`, `right`, or `same` for describing these relationships.\\

Your response should be in the format of:\\
<think><observation>...</observation><reasoning>...</reasoning><prediction>...\\
</prediction></think><answer>...</answer>\\

e.g. <think><observation>The box is below the player and the target is below and to the left of the player.</observation><reasoning>I need to go down to push the box down to the target.</reasoning><prediction>The player will be above the box, the target will be to the left of the box.</prediction></think><answer>Down</answer>\\
-------------------\\
(N-th Turn, N>0)\\
-------------------\\
The new observation is:\\
<image>\\

Decide your next action.\\
You should take 1 action at a time.\\

First, you should first describe the current observation.
Second, using this information, infer which action could plausibly have led to the current observation. Ignore and do not rely on your previous answer or intended action. Your inference must be based only on the observed state and the state transition from the previous step. 
Then provide your reasoning, predict the next state, and finally give your answer.\\

1) For the content you provide within the `<observation>` and `<prediction>` tags, you must strictly describe the relative position of the `target` and any visible `box` objects **relative to the player**. Your description/prediction must include **both** a vertical and a horizontal directional relationship for each object. Use ONLY the terms `above`, `below`, `left`, `right`, or `same` for describing these relationships. \\
2) For the content you provide within the `<retrospection>`, you must strictly assign a score to each action in [Up, Down, Left, Right]. Output a vector of length 4, with integer scores ranging from -5 to 5. A higher score indicates a higher likelihood that the action was taken to reach the current observation.\\

Your response should be in the format of:\\
<think><observation>...</observation><retrospection>...</retrospection>\\
<reasoning>...</reasoning><prediction>...</prediction></think>\\
<answer>...</answer>\\

e.g. <think><observation>The box is below the player and the target is below and to the left of the player.</observation><retrospection>I think the scores of actions in [Up, Down, Left, Right] for causing the environment state transition is: [0,5,-1,-5].</retrospection><reasoning>I need to go down to push the box down to the target.</reasoning><prediction>The player will be above the box, the target will be to the left of the box.</prediction></think><answer>Down</answer>\\
\end{tcolorbox}

\captionof{table}{Prompt for FrozenLake.}
\label{prompt:prompt_frozenlake}
\begin{tcolorbox}[
  enhanced,                 
  breakable,                
  width=\linewidth,
  colback=gray!10,
  colframe=gray!50,
  boxrule=0.4pt,
  arc=2pt,
  left=6pt,
  right=6pt,
  top=6pt,
  bottom=6pt,
  fonttitle=\itshape\small\color{black},
  fontupper=\small\ttfamily
]
\ttfamily
\textbf{Prompt for FrozenLake}\\
----------------------\\
(Initial Turn)\\
----------------------\\
You are a FrozenLake solver.\\
Your task is to reach the goal and avoid falling into holes.\\

Actions you can take: Left, Down, Right, Up. \\
You should take 1 action at a time.\\

Initial Observation:\\
<image>\\

Decide your next action.\\
You should take 1 action at a time.\\

You should first describe the observation, then your reasoning, then predict the next state, and finally your answer.\\

For the content you provide within the `<observation>` and `<prediction>` tags, you must strictly describe the relative position of the `target` (the gift box) and any visible `hole` (blue circles) objects **relative to the player**. Your description/prediction must include **both** a vertical and a horizontal directional relationship for each object. Use ONLY the terms `above`, `below`, `left`, `right`, or `same` for describing these relationships.\\

Your response should be in the format of:\\
<think><observation>...</observation><reasoning>...</reasoning><prediction>...\\
</prediction></think><answer>...</answer>\\

e.g. <think><observation>The player is above and on the right side of target. There is a hole below and at the left of the player.</observation><reasoning>I should go down first.</reasoning><prediction>The player will be above and on the right side of target. There is a hole at the left of the player.</prediction></think><answer>Down</answer>\\
----------------------\\
(N-th Turn, N>0)\\
----------------------\\
The new observation is:\\
<image>\\

Decide your next action.\\
You should take 1 action at a time.\\

First, you should first describe the current observation.
Second, using this information, infer which action could plausibly have led to the current observation. Ignore and do not rely on your previous answer or intended action. Your inference must be based only on the observed state and the state transition from the previous step. 
Then provide your reasoning, predict the next state, and finally give your answer.\\

1) For the content you provide within the `<observation>` and `<prediction>` tags, you must strictly describe the relative position of the `target` and any visible `box` objects **relative to the player**. Your description/prediction must include **both** a vertical and a horizontal directional relationship for each object. Use ONLY the terms `above`, `below`, `left`, `right`, or `same` for describing these relationships.  \\
2) For the content you provide within the `<retrospection>`, you must strictly assign a score to each action in [Left, Down, Right, Up]. Output a vector of length 4, with integer scores ranging from -5 to 5. A higher score indicates a higher likelihood that the action was taken to reach the current observation.\\

Your response should be in the format of:\\
<think><observation>...</observation><retrospection>...</retrospection>\\
<reasoning>...</reasoning><prediction>...</prediction></think>\\
<answer>...</answer>\\

e.g. <think><observation>The player is on the right side of the goal. There is a hole at the left of the player.</observation><retrospection>I think the scores of actions in [Left, Down, Right, Up] for causing the environment state transition is: [0,5,-1,-5].</retrospection><reasoning>I should go down first.</reasoning><prediction>The player will be on the right side of the goal. There is a hole at the left of the player.</prediction></think><answer>Down</answer>\\
\end{tcolorbox}

\captionof{table}{Prompt for Navigation.}
\label{prompt:prompt_navigation}
\begin{tcolorbox}[
  enhanced,                 
  breakable,                
  width=\linewidth,
  colback=gray!10,
  colframe=gray!50,
  boxrule=0.4pt,
  arc=2pt,
  left=6pt,
  right=6pt,
  top=6pt,
  bottom=6pt,
  fonttitle=\itshape\small\color{black},
  fontupper=\small\ttfamily
]
\ttfamily
\textbf{Prompt for Navigation}\\
----------------------\\
(Initial Turn)\\
----------------------\\
You are a home robot and perform navigation tasks according to instructions.\\

Actions you can take: moveahead, moveback, moveright, moveleft, rotateright, rotateleft, lookup, lookdown. \\

The instruction will be provided with each observation. Look at the image carefully and navigate to complete the instruction.\\

Rules:\\
1. You should take 1 action at a time.\\
2. If you find yourself seems to be stuck, you can lookdown or lookup to see if there's any object above or below you, you can also rotate to see if there's any object behind you.\\

Initial Observation:\\
<image> \\

Human Instruction: \\
I am looking for a garbage can. Can you navigate to that object? \\

Decide your next action. \\
You can take 1 action at a time. \\

You should first give your thought process with the your observation, reasoning, and prediction of next state, then your answer. Both the observation and prediction should describe what you see or expect to see in the environment.\\

Your response should be in the format of:\\
<think><observation>...</observation><reasoning>...</reasoning><prediction>
...</prediction></think><answer>...</answer>\\

e.g. <think><observation>There is a garbage can in the upper left corner of the image.</observation><reasoning>To move to the garbage can, we can go forward-left, but since there's a kitchen counter directly ahead, we should go left first. </reasoning><prediction> I will be in front of the garbage can.</prediction></think>
<answer>moveleft</answer>\\
----------------------\\
(N-th Turn, N>0)\\
----------------------\\
The new observation is:\\
<image>\\

Decide your next action.\\
You should take 1 action at a time.\\

First, you should first describe the current observation.
Second, using this information, infer which action could plausibly have led to the current observation. Ignore and do not rely on your previous answer or intended action. Your inference must be based only on the observed state and the state transition from the previous step. 
Then provide your reasoning, predict the next state, and finally give your answer.\\

1) Both the observation and prediction should describe what you see or expect to see in the environment.\\
2) For the content you provide within the `<retrospection>`, you must strictly assign a score to each action in [moveahead, moveback, moveright, moveleft, rotateright, rotateleft, lookup, lookdown]. Output a vector of length 8, with integer scores ranging from -5 to 5. A higher score indicates a higher likelihood that the action was taken to reach the current observation.\\

Your response should be in the format of:\\
<think><observation>...</observation><retrospection>...</retrospection>\\
<reasoning>...</reasoning><prediction>...</prediction></think>\\
<answer>...</answer>\\

e.g. <think><observation>There is a garbage can in the upper left corner of the image. </observation> <retrospection> I think the scores of actions in [moveahead, moveback, moveright, moveleft, rotateright, rotateleft, lookup, lookdown] for causing the environment state transition is: [0,-5,-1,5,-5,-5,-5,-5].</retrospection><reasoning> To move to the garbage can, we can go forward-left, but since there's a kitchen counter directly ahead, we should go left first.</reasoning><prediction> I will be in front of the garbage can.</prediction></think><answer>moveleft</answer>\\
\end{tcolorbox}

\end{document}